\documentclass[10pt,twocolumn,letterpaper]{article}
\usepackage{iccv}  
\definecolor{iccvblue}{rgb}{0.21,0.49,0.74}
\usepackage[pagebackref,breaklinks,colorlinks,allcolors=iccvblue]{hyperref}

\title{DreamInsert: Zero-Shot Image-to-Video Object Insertion from A Single Image}
	
\author{
  Qi Zhao\textsuperscript{1,2} \quad
  Zhan Ma\textsuperscript{1} \quad
  Pan Zhou\textsuperscript{2} \quad \\[4pt]
  \textsuperscript{1}Nanjing University \quad \textsuperscript{2}Singapore Management University \\[4pt]
  \texttt{qizhao@smail.nju.edu.cn} \quad
\texttt{mazhan@nju.edu.cn} \quad
\texttt{panzhou@smu.edu.sg} \\ 
}

\begin{document}

\twocolumn[{%
\renewcommand\twocolumn[1][]{#1}%
\maketitle
\begin{center}
  \centering
   \setlength\belowdisplayskip{-0.1in}
   \includegraphics[width=0.87\linewidth]{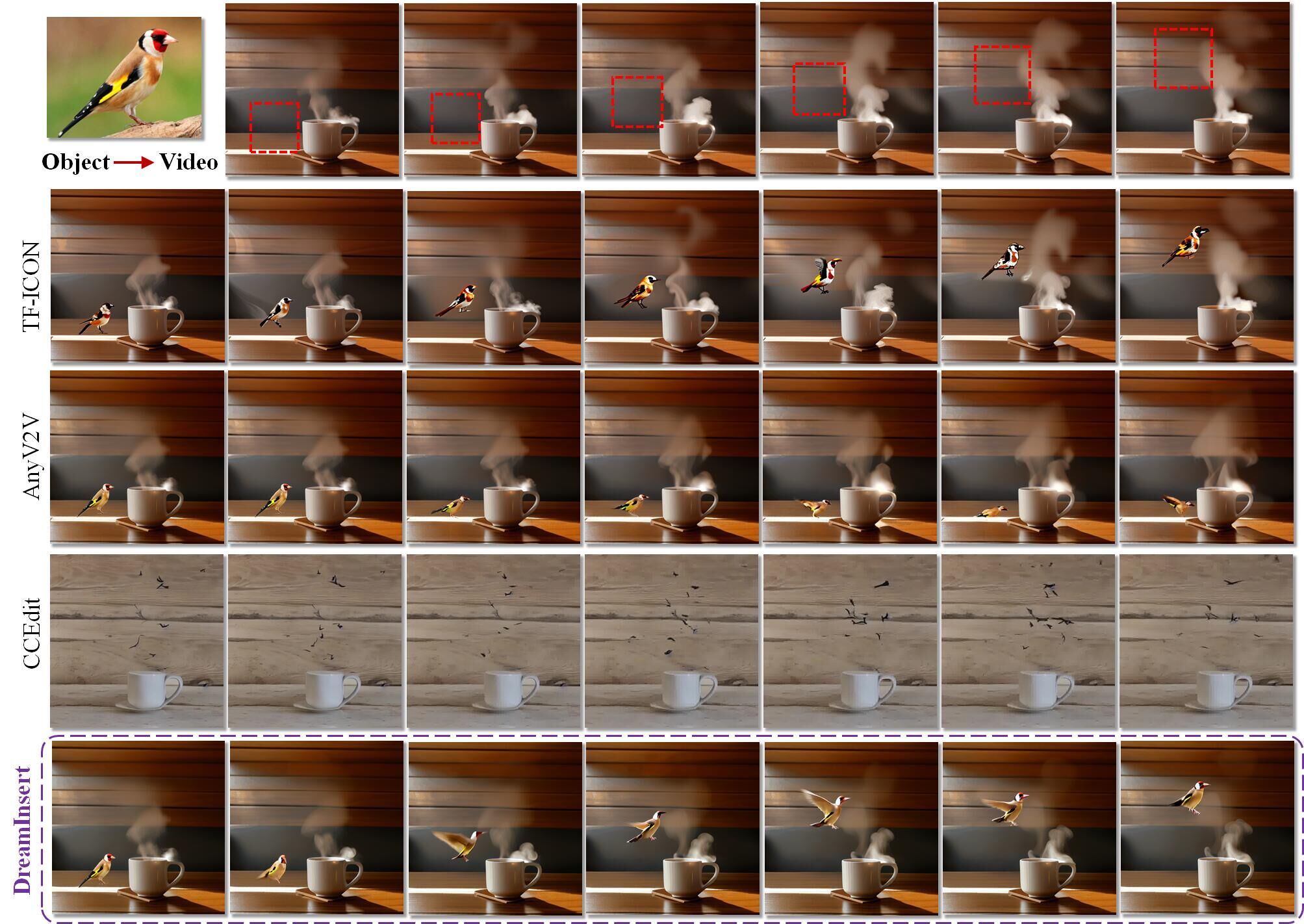}
    \setlength{\belowcaptionskip}{0.05in}
    \setlength{\abovecaptionskip}{0.05in}
   \captionof{figure}{
   Visual examples on ``Coffee-Bird'', where DreamInsert realizes zero-shot insertion for static object into dynamic video.}
   \label{abs_show}
\end{center}
}]
\vspace{-0.05in}

\maketitle
\begin{abstract}
Recent developments in generative diffusion models have turned many dreams into realities. For video object insertion, existing methods typically require additional information, such as a reference video or a 3D asset of the object, to generate the synthetic motion. However, inserting an object from a single reference photo into a target background video remains an uncharted area due to the lack of unseen motion information.
We propose DreamInsert, which achieves Image-to-Video Object Insertion in a training-free manner for the first time. 
By incorporating the trajectory of the object into consideration, DreamInsert can predict the unseen object movement, fuse it harmoniously with the background video, and generate the desired video seamlessly. More significantly, DreamInsert is both simple and effective, achieving zero-shot insertion without end-to-end training or additional fine-tuning on well-designed image-video data pairs.
We demonstrated the effectiveness of DreamInsert through a variety of experiments, examples of which are shown in Fig.~\ref{abs_show}. Leveraging this capability, we present the first results for Image-to-Video object insertion in a training-free manner, paving exciting new directions for future content creation and synthesis. 
The code will be released soon.
\end{abstract}
\vspace{-0.05in}    
\section{Introduction}
\label{sec:intro}
Memory and imagination are two fundamental functions of the human brain that play a crucial role in how we perceive our surroundings~\cite{mmpf}. Humans can recall vivid scenes and reshape memories with imagined alterations, reflects the entanglement between memory and imagination. 
In practice, inserting an object into a video is a challenging task with applications in various fields such as film production~\cite{ccdm}, virtual reality~\cite{vrp}, and therapeutic visualizations~\cite{seeme}. 

With recent progress in diffusion model~\cite{icml15, ddpm, ddim, flux}, such as text conditional generation~\cite{imagen, sdxl, sd3, videoldm, makeavideo}, image-to-video (I2V) generation~\cite{I2VEdit, i2vgenxl}, and image/video editing~\cite{controlnet, dreambooth, tficon}, object insertion has shown impressive results across different scenarios, such as image-to-image (I2I)~\cite{phdiffusion, tficon}, video-to-video (V2V)~\cite{IVV, aas}, and object-to-3D (O23D)~\cite{Instruct-NeRF2NeRF, Dipir, inserf}.  
Each scenario presents unique challenges, such as ensuring multiview consistency in O23D or harmonizing style attributes like lighting and shadow in I2I tasks.

Despite these advances, \textit{Image-to-Video (I2V) object insertion} remains an unexplored frontier. Formally, given a static object from a single image, how can we inject it into a dynamic video that aligns with background within desired movements? This I2V insertion task is particularly valuable for applications in fields like virtual reality and content creation~\cite{ldm, motionctrl, Dipir}, where realistic and adaptable object placements can greatly enhance user experience. 

\noindent{\textbf{Challenges.}}
The task of I2V insertion presents significant challenges in two key aspects.  \textit{First}, integrating a static object from an image into a dynamic video requires generating realistic motion that aligns with the object's natural behavior. For instance, a dog can run but cannot fly. Unlike I2I  and O23D  which do not impose constraints on motion and V2V that benefits from a reference motion, I2V insertion is inherently ill-posed. The absence of a direct motion reference creates ambiguity—there are infinitely many plausible ways an object might behave within a video.  

\noindent\textit{Challenge 1: Lack of reference motion when inserting a static object into a dynamic video.}  

\textit{Second}, even after generating motion for the inserted object, maintaining both spatial and temporal consistency across video frames is a formidable challenge. When a static object is introduced into a background video, discrepancies in movement, positioning, and object-background interactions can cause unnatural distortions. Ensuring the seamless integration of the object while preserving environmental coherence is a key differentiator between I2V insertion and other video synthesis tasks like I2V or T2V generation.  More critically, modeling optimal object-environment interactions through supervised learning or fine-tuning is nearly infeasible. Data pairs depicting the same object exhibiting various behaviors in identical scenes are virtually non-existent. Furthermore, training on imbalanced object-video distributions risks severe mode collapse~\cite{mode_c}, limiting the model’s ability to generalize across diverse scenarios.  

\noindent\textit{Challenge 2: The generated object motion must maintain both spatiotemporal and environmental consistency.}  

\noindent{\textbf{Contributions}}.  To address these challenges, we introduce \textbf{DreamInsert}, a novel zero-shot I2V insertion framework that enables controllable and training-free object integration into videos via a two-stage process.   To tackle Challenge 1, DreamInsert leverages textual descriptions and a predefined  trajectory sequences  as conditioning inputs to generate reasonable but coarse  motion for the static object in the first stage.   For Challenge 2, DreamInsert refines the generated motions in the second stage, ensuring spatiotemporal consistency and seamless environmental integration under the guidance of a pretrained generative model.  By adopting a training-free paradigm, DreamInsert circumvents the need for costly model fine-tuning, enabling flexible and adaptive object insertion without requiring additional training data. Experimental results validate its effectiveness across a wide range of scenarios. For instance, visual examples of the ``Coffee-Bird"   are illustrated in Fig.~\ref{abs_show}.  

\begin{itemize}
	\item[-] We introduce DreamInsert as the first training-free solution for I2V insertion, incorporating textual descriptions and a trajectory sequences as conditioning signals to generate realistic and non-existent object motions.  
	\item[-] Leveraging pretrained knowledge, DreamInsert achieves plausible object insertions without additional fine-tuning, effectively addressing the challenge of insufficient training data pairs.  
	\item[-] Through various experiments, we demonstrate DreamInsert's ability to insert static objects into videos with diverse backgrounds while maintaining natural and coherent object-environment interactions.  
\end{itemize}
\section{Related work}
\label{sec:formatting}
\noindent{\textbf{Controllable Video Generation.}} 
The stability and efficiency of diffusion models~\cite{icml15, ddpm, ddim} in density estimation have accelerated advances in content generation~\cite{gan, vqvae, vqgan, stylegan, ldm, glide, gd, sdxl, vdm, svd}, enabling controllable video generation~\cite{videoldm, higen, Lumiere, dit, sit, videolcm, makeavideo, latte, Peekaboo, CogVideo, Imagen-v, walt, Instantbooth, controlnet, controlnext, dreamfusion, dreamvideo2, animatediff, motionclone, motionctrl, cameractrl, ipadapter, videobooth, videocomposer, videodreamer, VideoSwap, turnavideo, controlvideo, CustomVideo, FancyVideo, video-p2p, videocrafter1, videocrafter2, dreamvideo, dreambooth, i2vgenxl, i2v-adapter, I2VEdit, motioni2v, makeitmove}. For controllable video generation, some researchers fine-tune text-to-image (T2I) models~\cite{ldm, sdxl} on multimodal data pairs~\cite{videoldm, vdm, makeavideo} or customized datasets~\cite{dreambooth, animatediff}, enabling Text-to-Video (T2V) generation. Other studies use motion vectors~\cite{videocomposer}, Plücker embeddings for camera paths~\cite{cameractrl}, and joint motion for camera-object relations~\cite{motionctrl} as condition signals for fine-grained control. Besides, subject-driven generation~\cite{dreambooth, dreamtuner, videobooth, jedi, suti} is gradually emerging as a promising direction, aiming to generate consistent content tailored to specific subjects.

\noindent{\textbf{Training-free Content Editing.}} 
Some works fine-tune the backbone model for content editing~\cite{text2live, gen1, video-p2p, ccedit, sdedit, catvton, controlnet}, but may harm model's generalization ability. Training-free methods offer a compelling alternative that uses noise inversion and feature injection for content editing~\cite{text2live, gen1, video-p2p, ccedit, sdedit, I2VEdit, catvton, anyv2v, uniedit, SVC, mvoc, ptp, pnp, PrMAc, null, textinv, infedit}. PtP~\cite{ptp} discovers the importance of cross-attention layers to control the relation between the image layout and words in the prompt.
PnP~\cite{pnp} observes that the control of the generated content can be achieved by manipulating the spatial features and their self-attention. 
AnyV2V~\cite{anyv2v} divides the video editing task into first frame editing and I2V generation, leading to controllable video composition.		
 
Some works aim to address the mismatch during inversion processes. Classifier-Free Guidance (CFG)~\cite{cfg} has been shown to have impacts on content fidelity~\cite{null, tficon, edict}. 
SDEdit-based methods~\cite{sdedit, eie, gradop, p2v} apply perturbations to pixels before inversion. Precise control over space and time remains a challenge in content editing.
 
\noindent{\textbf{Object Insertion.}} 
Research on object insertion has focused on image-based (I2I) and 3D scene insertion. For videos, the focus has been on rigid body insertion and light rendering, with limited work on predicting object dynamics. I2I methods~\cite{dovenet, phdiffusion, relh, magici} often employ inpainting models to learn interactions between foreground and background. Training-free image composition approaches, such as~\cite{tficon}, leverage powerful T2I diffusion models for object blending.
 
3D object insertion~\cite{Instruct-NeRF2NeRF, RGCI, FocalDreamer, inserf, Dipir} involves integrating objects into 3D scenes, especially using Neural Radiance Fields (NeRF).
Instruct-NeRF2NeRF~\cite{Instruct-NeRF2NeRF} first uses text guidance for 3D editing and InseRF~\cite{inserf} achieves rough object placement from single views. The main challenges in 3D insertion include multiview consistency and realistic rendering. In contrast, I2V insertion emphasizes generating unseen motion while maintaining consistency.
%\section{Method}
\section{Preliminary}\label{preliminary}
We introduce the preliminary of Latent Diffusion Model (LDM) and DDIM inversion which being pivotal in the DreamInsert, to better illustrate the proposed pipeline.

\noindent \textbf{Latent Diffusion Model}. LDM is used in models of both coarse injection and spatiotemporal alignment stages.
In LDM, input images $\mathbf{z}_0$ will be encoded in latents space as $z_0$ by VAE. This pixel compression process saves computational costs while maintaining semantic features, and also helps improve the diversity and editability of diffusion model generation~\cite{latentpaint, cads}. During forward process, the encoded latents $z_0$ will be perturbed by noise into $z_t$:
\[
z_t = \sqrt{\Bar{\alpha}_t}z_0 + \sqrt{1-\Bar{\alpha}_t}\epsilon, ~\epsilon \sim\mathcal{N}(0, I),  
\]
where $\Bar{\alpha}_t$ determines the noise strength according to step $t$. During the denoising process, the noise patterns will be modeled via a noise prediction model $\mathbf{s_\theta}$ with conditional signal $c$ by minimizing the training objective:
\[
\arg \min \mathbb{E}_{z_0, \epsilon \sim\mathcal{N}(0, I), t, c}[\Vert \epsilon - \mathbf{s_\theta}(z_0, t, c) \Vert].
\]
Once the training is completed, new images can be decoded by gradually denoising the samples from $p(z)$,
\begin{equation*}\small
z_{t-1} = \sqrt{\frac{\Bar{\alpha}_{t-1}} 
{\Bar{\alpha_{t} } } }  z_{t} + \left( \sqrt{\frac{1}{\Bar{\alpha}_{t-1}} -1} - \sqrt{\frac{1}{\Bar{\alpha}_{t}} -1}   \right) \cdot \mathbf{s_\theta}(z_t, t, c).
\end{equation*}

\noindent \textbf{Inversion-based Editing.} 
When we obtain the score model $\mathbf{s_\theta}$, we can achieve the mapping between noise and image through inversion. Specifically, we can use DDIM~\cite{ddim} inversion to obtain the corresponding noise latents,
\begin{equation*}\small
z_{t+1} = \sqrt{\frac{\Bar{\alpha}_{t+1}} 
{\Bar{\alpha_{t} } } }  z_{t} + \left( \sqrt{\frac{1}{\Bar{\alpha}_{t+1}} -1} - \sqrt{\frac{1}{\Bar{\alpha}_{t}} -1}   \right)  \cdot \mathbf{s_\theta}(z_t, t, c).
\end{equation*}
Once we obtain the inverted latent representation for the images or video frames, we can modify those content by manipulating the latents, adding new conditions, and feature injection, namely \textit{Inversion-based Editing}~\cite{null, pnp, ptp, sdedit, tf-t2v, ConFiner, anyv2v}.
\section{DreamInsert}
\begin{figure*}[!t]
  \centering
 \includegraphics[width=0.9\linewidth]{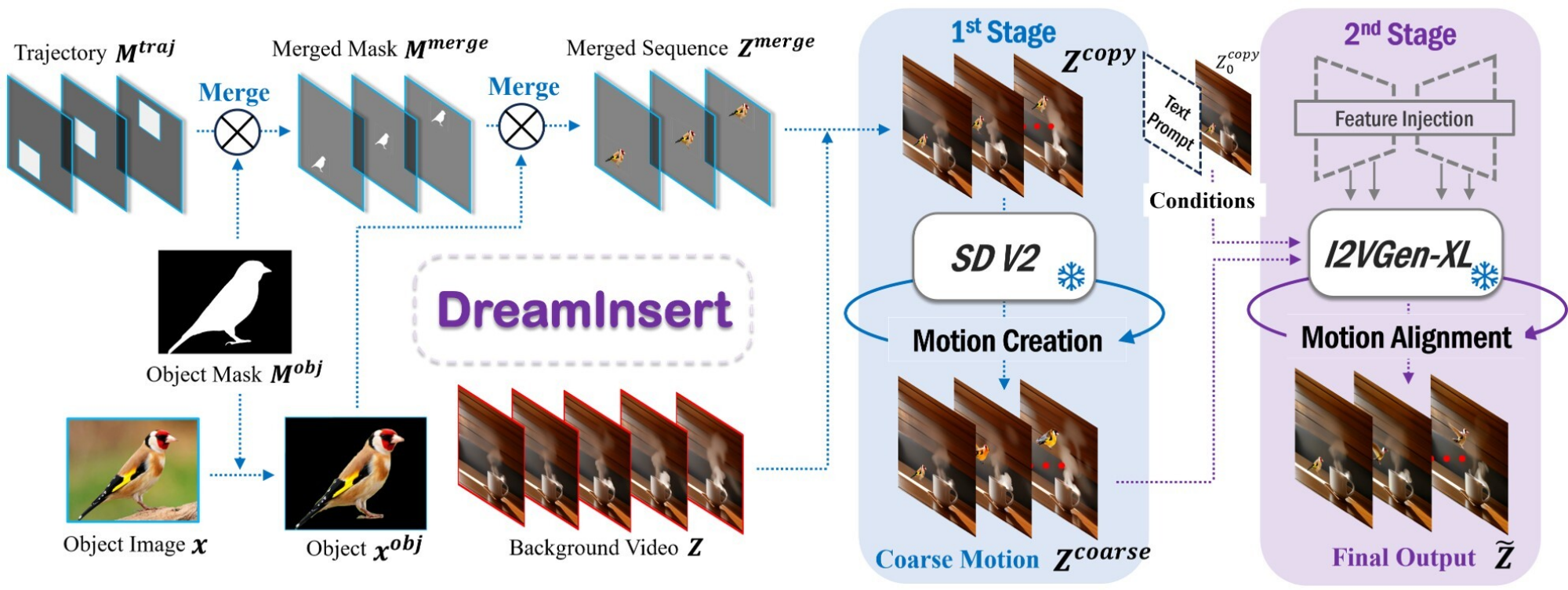}\\
  \caption{
  The overview of DreamInsert where $\mathtt{Merge}$ denotes the rescale + replace operation and $\mathbf{Z}^\text{merge} = \mathtt{Merge}(\mathbf{x}^{\text{obj}}, \mathbf{M}^{\text{merge}} )$, showcasing two stages: the \textcolor{blue}{blue} part is the first stage of motion creation, while the \textcolor{violet}{purple} part is the second stage of spatiotemporal alignment. 
  }
  \label{dreaminsert}%\vspace{-0.2in}
\end{figure*}

Fig.~\ref{dreaminsert} illustrates the overall framework of DreamInsert, structured as a two-stage process to systematically address the two challenges of I2V insertion.  Sec.~\ref{overview_framework} provides an overview of the framework which consists of two stages. The first stage, Motion Creation, is detailed in Sec.~\ref{d_inv}, while the second stage, Spatiotemporal Alignment, is described in Sec.~\ref{st_align}.  
 
\subsection{Overall Framework}\label{overview_framework}
As discussed earlier, I2V insertion faces two fundamental challenges: (1) the absence of reference motion for static objects and (2) the need for spatiotemporal consistency to ensure seamless integration into a dynamic video. To tackle these challenges, we decompose the object's motion into two hierarchical components: coarse- and fine-grained motions. The former  defines the overall movement direction and speed of the object while keeping the course-grained appearance of object, while the later ensures the realism of the object's appearance, maintains spatiotemporal coherence, and adapts to the surrounding environment.  

To generate coarse-grained motion, we leverage a predefined (bounding box level) movement trajectory  and textual descriptions as conditioning signals together with the static image to generate   a sequence of plausible coarse motion trajectories. This forms Stage 1 of DreamInsert, addressing Challenge 1.  Next, Stage 2 resolves Challenge 2 and  refines  coarse-grained motion into a fine-grained motion  via a pretrained I2V model, ensuring both object integrity and environmental consistency. The core motivation behind this two-stage design is that I2V insertion requires a delicate balance between preserving background fidelity and synthesizing realistic object motion—a challenge that cannot be effectively handled through one-shot generation.  

\subsection{First-stage: Motion Creation}\label{d_inv}
Here we study how to use object image $\mathbf{x}$, background video $\mathbf{Z}$, and trajectory sequence $\mathbf{M}^{\text{traj}}$ to generate a  coarse-grained motion of the object. This coarse-grained  motion can be obtained by a frame-wise merging operation and can be further refined in the subsequent stage for consistency. 

As shown in Fig.~\ref{dreaminsert}, for a static object in the reference image $\mathbf{x}$, we use a predefined bounding-box (bbox) sequence $\mathbf{M}^{\text{traj}}$ to control its movement trajectory, where   $\mathbf{M}^{\text{traj}}$ has the same sequence length as the background video $\mathbf{Z}$. Each frame in $\mathbf{M}^{\text{traj}}$ contains a bbox that indicates the expected position of the object, where pixels within the object area are set to 1 and the rest are set to 0. This sequence provides guidance for objects' movement.

To generate the coarse-grained motion for object by using $\mathbf{M}^{\text{traj}}$,  we use Segment Anything 2~\cite{sam2}, a segmentation model,  to   extract the object segmentation mask $\mathbf{M}^{\text{obj}}$   as shown in Fig.~\ref{dreaminsert}. Then we resize $\mathbf{M}^{\text{obj}}$ to fit the bbox in $\mathbf{M}^{\text{traj}}$ for replacement, yielding a frame-specific merged mask:
\[
\mathbf{M}^{\text{merge}}_i = \mathtt{Merge}(\mathbf{M}^{\text{obj}}, \mathbf{M}^{\text{traj}}_i), \quad i \in [0,N]
\]
where $\mathbf{M}^{\text{merge}}_i$ is the merged mask of the $i$-th frame. As shown in Fig.~\ref{dreaminsert}, $\mathtt{Merge}$ denotes the operation that resize the object mask $\mathbf{M}^{\text{obj}}$ to the same shape of the bbox in $\mathbf{M}^{\text{traj}}_i$ and then replace it, forming a merged mask.  
Consequently, we first segment the object $\mathbf{x}^\text{obj}$ from the image $\mathbf{x}$ using $\mathbf{M}^{\text{obj}}$, $\mathbf{x}^\text{obj} = \mathbf{x} \odot \mathbf{M}^{\text{obj}}$, where $\odot$ denotes the Hadamard product, i.e., element-wise product. Then, we resize and copy the $\mathbf{x}^\text{obj}$ according to $\mathbf{M}^{\text{merge}}_i$ into the corresponding background frame $\mathbf{Z}_i$ and obtain $\mathbf{Z}^{\text{copy}}_i$,
\begin{equation}\label{z_copy}\small 
\mathbf{Z}^{\text{copy}}_i = \mathtt{Merge}(\mathbf{x}^{\text{obj}}, \mathbf{M}^{\text{merge}}_i ) + \mathbf{Z}_i \odot (1 - \mathbf{M}^{\text{merge}}_i). \nonumber
\end{equation}
Then to create coarse-grained motion, we explore two approaches: pixel noise injection and latent noise injection. While the former is straightforward and computationally efficient, the latter, which leverages diffusion inversion, often delivers superior results as shown in Sec.~\ref{exp}.  

\begin{figure}[!t]
	\centering
\includegraphics[width=1\linewidth]{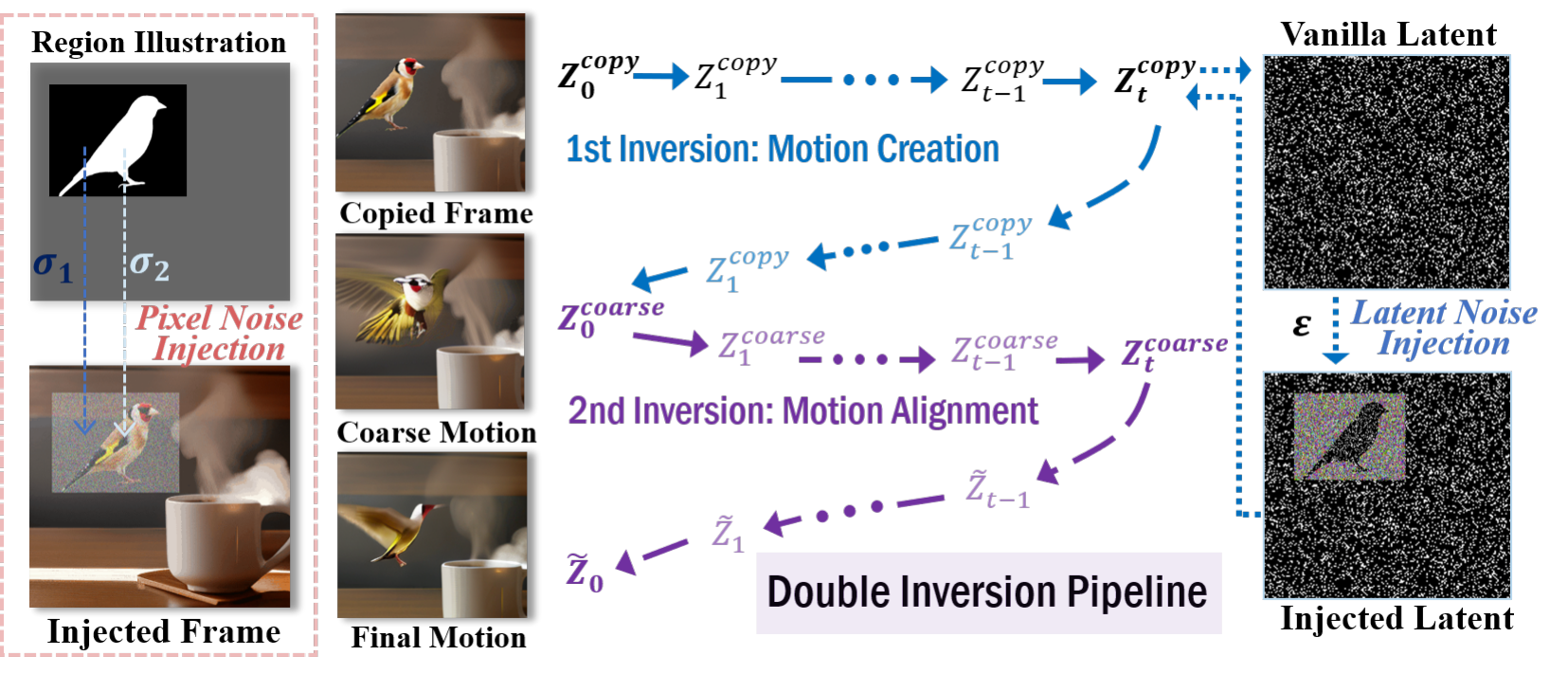}\\
	\caption{
		\textit{Left}: Pixel noise injection with the region illustration, where gray for the background, white for the object area and black for the interaction (IA) area. \textit{Right}: The overview of Double Inversion pipeline with latent noise injection. We only add noise in the latent's IA area and obtain coarse frame after denoising.}
	\label{d_inv_overview}
\end{figure}

\noindent{\textbf{Pixel Noise Injection}.}\label{pn_inj} 
To generate motion for static objects efficiently, we introduce a lightweight Pixel Noise Injection (PN-Inj) that selectively perturbs key regions while preserving the background. We obtain the coarse motion sequence $\mathbf{Z}^{\text{coarse}}$ by adding less Gaussian noise $\epsilon \sim \mathcal{N}(0,1)$ to the object and more Gaussian noise into  its interaction areas in each video frame, while keeping the background area fixed. A visual representation of this region division is shown in Fig.~\ref{d_inv_overview} (left).  For each frame $\mathbf{Z}^{\text{copy}}_i$, we apply different noise intensities to different regions:
\begin{equation*}
	\label{pninj}
	\fontsize{8}{3}\selectfont{
\begin{split} 
	& \mathbf{Z}^{\text{coarse}}_i =   \mathbf{Z}^{\text{copy}}_i \odot (1 - \mathbf{M}^{\text{traj}}_i)  \qquad \qquad \qquad \quad \ \ \ \text{(background: no noise)} \\
	& \quad +  \sigma_1 * \epsilon \odot \mathbf{M}^{\text{IA}}_i + (1-\sigma_1) * \mathbf{Z}^{\text{copy}}_i \odot \mathbf{M}^{\text{IA}}_i \ \  \text{(interaction: more noise)} \\
	& \quad+ \sigma_2 * \epsilon \odot \mathbf{M}^{\text{merge}}_i + (1-\sigma_2) * \mathbf{Z}^{\text{copy}}_i \odot \mathbf{M}^{\text{merge}}_i,    \text{(object: less noise)}  
	\end{split}
}
\end{equation*} 
where $\oplus$ is XOR operation and $\mathbf{M}^{\text{IA}}_i = \mathbf{M}^{\text{merge}}_i \oplus \mathbf{M}^{\text{traj}}_i$ denotes the interaction (IA) area as shown in Fig.~\ref{d_inv_overview} (left). $\sigma_1$ and $\sigma_2$  control  noise intensity of interaction and object area, respectively.

After applying different noise injection strategies, in the second stage we will refine the coarse video $\mathbf{Z}^{\text{coarse}}$ using a text-to-video diffusion model. Similar to image editing techniques~\cite{pnp, ptp, sdedit}, injecting noise into specific regions of a video frame enables targeted modifications based on input conditions, such as textual descriptions in this work. Additionally, higher noise levels provide greater flexibility for editing.  
Given the image-to-video nature of our task and the need for controlled editing, we apply different noise intensities to distinct regions. First, since the object is initially static in each frame $\mathbf{Z}^{\text{copy}}_i$, it must be modified to exhibit motion while preserving its appearance, requiring noise injection for editing. Second, the interaction region between the object and the scene requires significant editing and thus needs stronger noise inejction, as inserting a new object alters occlusion, lighting, and shadows. Lastly, the background should remain unchanged to maintain visual fidelity, and thus should get rid of noise injection. By following this strategy, lower noise levels facilitate natural object motion generation, higher noise levels allow for more extensive interaction modifications, and the absence of noise in the background ensures consistency with the original scene.

\noindent{\textbf{Latent Noise Injection}.} 
While Pixel Noise Injection (PN-Inj) efficiently generates a coarse-grained motion sequence that captures the approximate appearance and position of an object, real-world applications often require finer control guided by textual descriptions. To address this, we introduce latent noise injection (LN-Inj) which injects noise into the latent representation of video inversion, enabling the generation of more natural object interactions and motion under text guidance during the denoising process of a text-to-image model.  

To achieve this, we employ inversion-based editing to incorporate the textual condition $\mathcal{C}^{\text{obj}}$ into $\mathbf{Z}^{\text{copy}}$ for coarse-grained motion generation. Specifically, we obtain the inverted noise representation, a.k.a., latent representation, by using SDv2~\cite{ldm}, an Ordinary Differential Equation (ODE)-based inversion method with DPM++~\cite{dpm} for text-to-image generation:  
\begin{equation}\label{inversion}
	\xi_i = \mathtt{Inv}_{I}(\mathbf{Z}^{\text{copy}}_i).
\end{equation}  
This latent representation  $\xi_i$ allow the reconstruction of the original input video frames through multiple denoising steps in SDv2 (see Sec.~\ref{preliminary}). Moreover, same as image editing techniques~\cite{pnp, ptp, sdedit}, injecting noise into specific regions of the video frame enables targeted modifications, ensuring that the edited regions align with the input conditions, such as the provided text description.   
 
Since the newly inserted object does not naturally blend with the  original background frame, we introduce Gaussian noise $\epsilon \sim \mathcal{N}(0,1)$ into the interaction area (IA) of the latent representations $\xi_i$. This  encourages the model to synthesize harmonious interactions from a new starting point $\hat{\xi}_i$:  
\[
\hat{\xi}_i = \xi_i \odot (1- \mathbf{m}^{\text{IA}}_i) + \epsilon \odot \mathbf{m}^{\text{IA}}_i,
\]  
where  $\mathbf{m}^{\text{IA}}_i$ represents the rescaled mask of $\mathbf{M}^{\text{IA}}_i$ in the latent space, and $\mathbf{M}^{\text{IA}}_i$ is the interaction area in the original frame space. Fig.~\ref{d_inv_overview} visualizes  this process, where   ``Injected Latent'' is the $\hat{\xi}_i$ and the colored region is the $\mathbf{m}^{\text{IA}}_i$. 

Finally, the refined frame $\mathbf{Z}^{\text{coarse}}_i$, which includes improved motion synthesis, is generated using the decoder $\mathtt{Dec}_I$ of the text-to-image (T2I) model, guided by the textual description $\mathcal{C}^{\text{obj}}$:  
\[
\mathbf{Z}^{\text{coarse}}_i = \mathtt{Dec}_{I}(  \{\hat{\xi}_i\}, \mathcal{C}^{\text{obj}}),  \quad i \in [0,N].
\]  
Compared with pixel noise injection,  this latent noise injection uses  the extra textual description $\mathcal{C}^{\text{obj}}$, and often provides better coarse-grained motion and also the final image-to-video insertion performance as shown in Sec.~\ref{exp}. 

\begin{figure}[!t]
	\centering
	\includegraphics[width=1\linewidth]{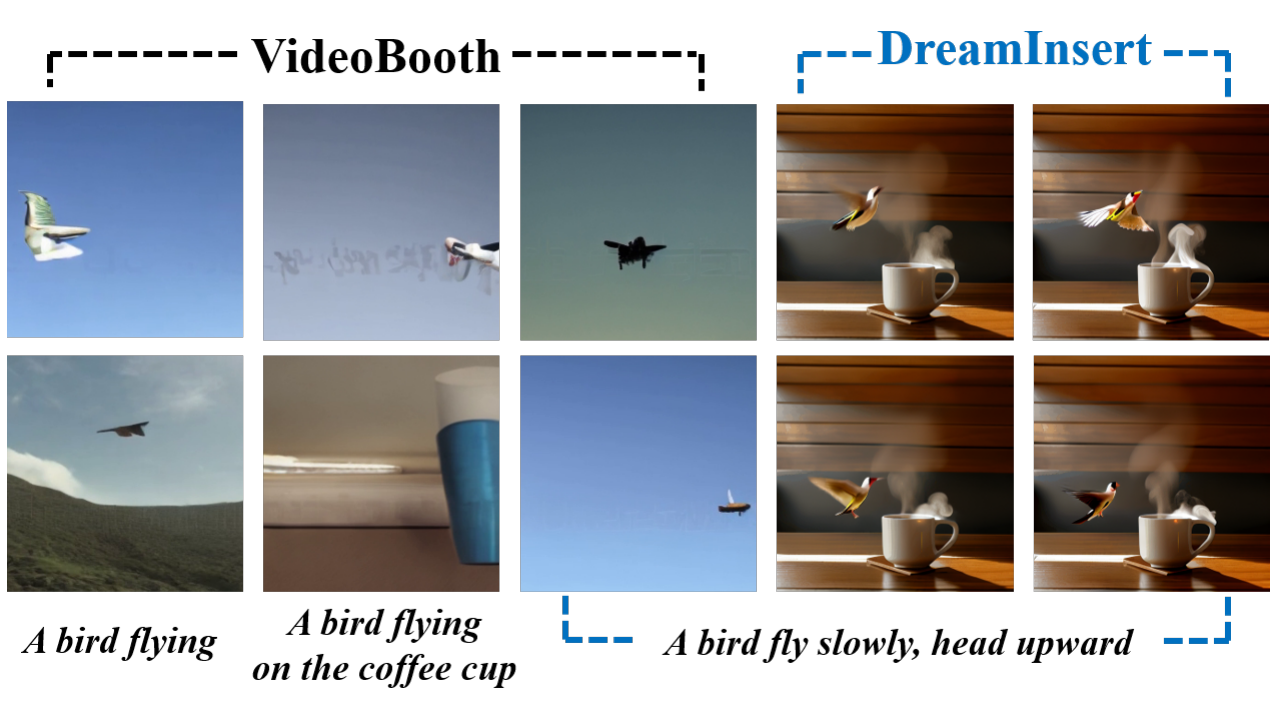}\\
	\caption{
		Existing training-based subject-driven method can hardly maintain consistency in scenarios of disordering semantic.
	}
	\label{booth}\vspace{-0.05in}
\end{figure}
\begin{figure*}[!t]
	\centering
	\includegraphics[width=0.9\linewidth]{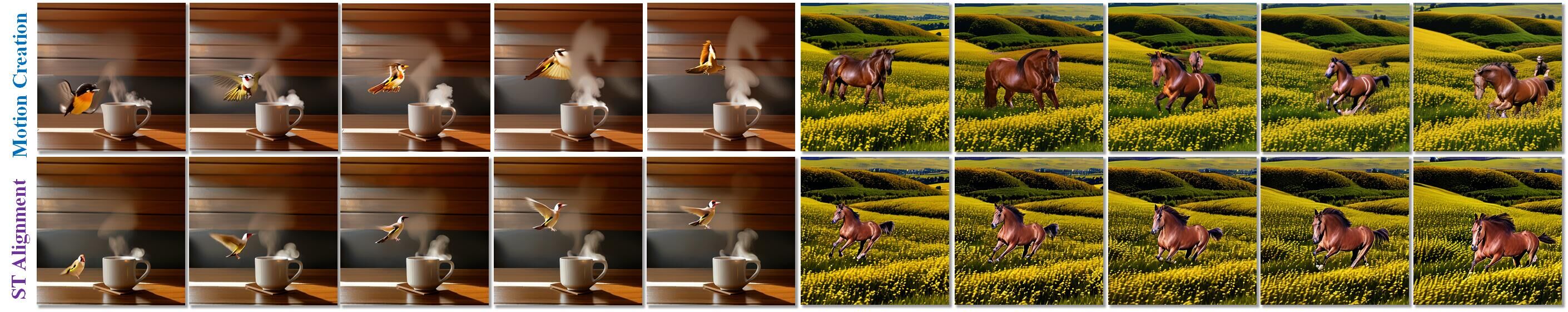}\\
	\caption{ 
		Visual examples of outputs in two stages. \textit{Top}: motion creation in the 1st stage; \textit{Bottom}: alignment in the 2nd stage.
	}
	\label{stage_abla} 
\end{figure*}\vspace{-0.05in}

\subsection{Second-stage: Spatiotemporal Alignment}\label{st_align}
The first stage generates a coarse video, \(\mathbf{Z}^{\text{coarse}}\), which provides an initial motion sequence for the inserted object. However, \(\mathbf{Z}^{\text{coarse}}\) often exhibits inconsistencies in both object appearance and its interaction with the background across frames, such as unnatural bird flight motions, as illustrated in Fig.~\ref{stage_abla} (top). To address this, the second stage refines \(\mathbf{Z}^{\text{coarse}}\) for  spatiotemporal coherence while preserving realistic object-background interactions.  
 
Moreover, such disordering semantic cases are rare in real-world datasets, making it infeasible to fine-tune backbone models for improved performance. Subject-driven adaptation methods~\cite{dreambooth, dreamtuner, videobooth, jedi, suti} struggle with this limitation, failing to maintain both fidelity and consistency. Fig.~\ref{booth} visualizes the inconsistencies produced by VideoBooth~\cite{videobooth} in such scenarios.

\noindent\textbf{Challenges for Spatiotemporal  Consistency.}  
Existing video  generation models~\cite{svd, Imagen-v, turnavideo, videobooth, i2vgenxl} are typically trained on large-scale text-image~\cite{laion400m, laion5b} or text-video datasets~\cite{svd, viclip, webvid}, allowing them to learn general semantic relationships between text prompts and natural scenes.  However, due to the ill-posed nature of image-to-video (I2V) object insertion, the synthesized video often deviates significantly from the videos in the training dataset. The inserted object may appear in unexpected contexts (e.g., a bird in an indoor scene with a coffee cup and table, as shown in Fig.~\ref{booth}), leading to rare and even novel scenarios. This semantic mismatch is a primary cause of spatiotemporal inconsistencies, making it difficult for existing models to generate fine-grained motion which is Challenge 2 as discussed in  Sec.~\ref{sec:intro}.   
Moreover, those  rare and even novel video scenarios   are hard to collect in reality and may not exist, making it challenging to finetune the backbone model to generate fine-grained motion. Owing to the lack of data, existing subject-driven methods~\cite{dreambooth, dreamtuner, videobooth, jedi, suti} can hardly handle this situation, unable to maintain fidelity and consistency. Different results using VideoBooth~\cite{videobooth} are visualized in Fig.~\ref{booth}.

\noindent{\textbf{Training-Free Editing}.} 
To overcome these challenges, we leverage the pretrained generative model’s spatiotemporal knowledge in a training-free manner. Our approach decouples the synthesis process into two steps. The background is reconstructed using the inverted latents of the coarse video, mitigating semantic mismatches between the inserted object and the scene.  The textual prompt is used exclusively to refine the object's motion while ensuring seamless integration during the denoising process.  

Given the coarse motion sequence, the copied object from the first frame, and a textual description, DreamInsert performs inversion-based editing to guide the denoising process toward a high-fidelity, temporally coherent output. Specifically,    
the coarse video $\mathbf{Z}^{\text{coarse}}$ is first inverted into the latent space of an I2V model via DDIM inversion~\cite{ddim}:
\begin{equation}\label{i2v_inv}
	\zeta = \mathtt{Inv}_{V}(\mathbf{Z}^{\text{coarse}}),
\end{equation}
where $\zeta$ represents the latent representation. Next, using the first frame of the inserted object $\mathbf{Z}^\text{copy}_0$ and textual alignment prompt $\mathcal{C}^\text{align}$, the I2V model~\cite{i2vgenxl} 
performs controlled denoising to generate the spatiotemporally aligned video:
\begin{equation}\label{i2v_dec}
	\tilde{\mathbf{Z}} = \mathbf{Z}^{\text{align}} = \mathtt{Dec}_{V} (\zeta, \mathbf{Z}_0^\text{copy}, \mathcal{C}^{\text{align}}).
\end{equation}
This   ensures that the inserted object blends naturally with the background while maintaining temporal coherence.

\noindent{\textbf{Feature Injection}.}
 To further refine spatialtemporal alignment, we introduce feature and attention score injection during the early denoising steps inspired by~\cite{pnp, anyv2v}. These techniques preserve crucial motion cues and prevent drift during the refinement process. Specifically, during the DDIM inversion in the $2^{nd}$ stage,   the  attention score is injected performed in the first 5 steps out of the  total denoising step of 50 which {guides the reconstruction of the inserted object and  maintains spatial consistency}.  Moreover, we perform spatial feature injection in the  first 5 steps  to {preserve the vanilla coarse movement and also reinforce motion continuity}.  By injecting coarse movement information from the initial video, these techniques ensure the final output maintains both realism and fidelity while preserving the object's intended motion. 

\noindent{\textbf{Discussions}.}  As shown in Fig.~\ref{d_inv_overview}, we refer to this two-stage inversion-based refinement pipeline as Double Inversion (D-Inv), contrasting it with PN-Inj,  as it enables the insertion of static objects into dynamic videos through two-stage inversion-based editing.  Unlike prior methods, D-Inv achieves seamless I2V insertion without additional fine-tuning or training data, and shows  strong robustness across diverse object-background combinations. Accordingly, D-Inv is set as the default pipeline for DreamInsert.
\section{Experiments}\label{exp}
We conducted comprehensive experiments to verify the effectiveness of DreamInsert. 
The main results contain quantitative and qualitative evaluations in Sec.~\ref{sec_quan_eval} and Sec.~\ref{sec_qual_eval}.
User study is given in Sec.~\ref{us} and ablation study in Sec.~\ref{as}.
More details and results are demonstrated in the Appendix.

\begin{figure*}[!t]
  \centering
 \includegraphics[width=0.9\linewidth]{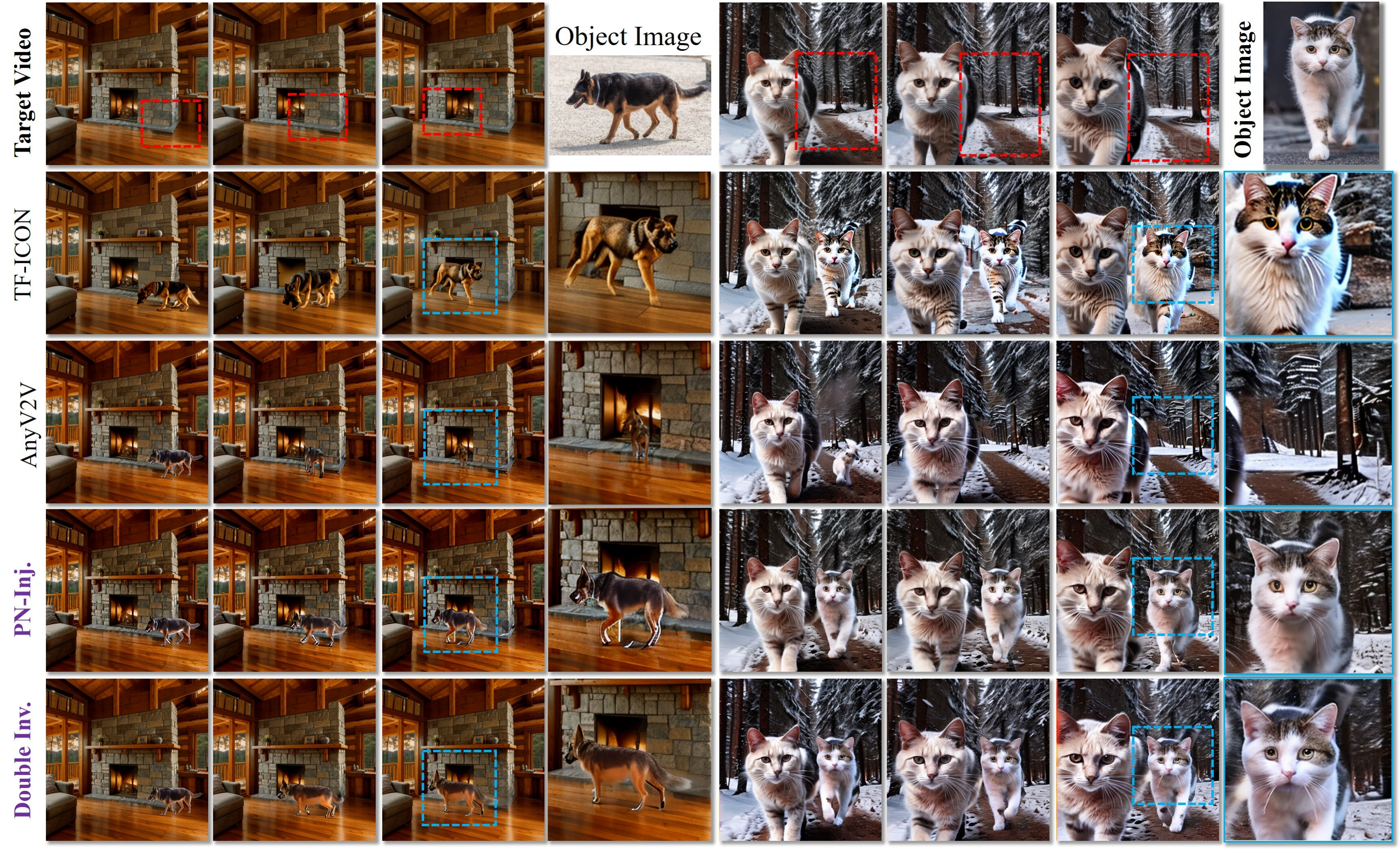}\\
  \caption{
  Visual comparison of DreamInsert with other methods, showing results at the 2nd, 8th, and 16th frames (from left to right).
  }
  \label{cat}\vspace{-0.05in}
\end{figure*}

\subsection{Settings.} 
To verify the performance of different methods on I2V insertion, we propose a dataset, I2V Insertion (I2VIns), and all experiments were conducted on I2VIns in an NVIDIA L40S GPU. 

\noindent \textbf{I2V Insertion Dataset}.
I2VIns dataset contains 14 object-video pairs used for evaluation, each case has a background video, an object image and a trajectory sequence. For background videos, six of them are from DAVIS16~\cite{davis16} and the others are generated by AnimateDiff~\cite{animatediff} using text prompts.
MotionLora~\cite{animatediff} is used to control camera movement in some of them and all prompts for the background videos please refer to the Appendix.
Each background video contains 16 frames with the shape $512 \times 512$, which covers various scenes from indoor to outdoor.
Objects' images are collected from the Internet.  
For each object-video pairs, we use a b-box sequence as the trajectory to control the movement of the object. 

\noindent \textbf{Metrics}. 
To quantitatively evaluate the synthesis results, we used different metrics. We use Clip-I~\cite{clip} and DINO~\cite{dino} score to evaluate the fidelity of insertion results, and Clip-T~\cite{clip} to evaluate the degree of text matching. In addition, we propose an Adv-Viclip score to evaluate the overall quality of the results.

\noindent \textbf{Baselines.} We select some of the latest content editing methods to compare with DreamInsert. \textit{TF-ICON}~\cite{tficon} is one of the SOTA image editing methods, and we use it for frame-wise object insertion. \textit{CCEdit}~\cite{ccedit} is a generalized video editing method trained on large-scale video datasets~\cite{webvid} using a T2I model as backbone. \textit{AnyV2V}~\cite{anyv2v} is a training-free video editing method that enables zero-shot editing of various downstream tasks. 

\subsection{Quantitative Evaluation}\label{sec_quan_eval}
\noindent \textbf{Fidelity.}
We use Clip-I score~\cite{clip} to calculate the similarity between the copied sequence and prediction to validate the \textit{background} fidelity. Besides, we use DINO score~\cite{dino} to calculate distance between object image and the b-box area of each synthesized frame to assess the \textit{object} fidelity. Results are shown in Tab.~\ref{quant}, DreamInsert acheves the best result of 91.77 in D-Inv setting.

\noindent \textbf{Text-Matching.} We calculate Clip Text (Clip-T)~\cite{clip} score between each model’s output and conditional prompt, thereby gauging the accuracy of the created motions instructed by the text. DreamInsert obtained the motion which is more in line with the prompt description.

\begin{table}[!h]\footnotesize
\centering
\setlength{\abovecaptionskip}{3pt}
\setlength{\belowcaptionskip}{-15pt} 
\tabcolsep=0.15cm
\begin{tabular}{l||ccc|cc}
  & AnyV2V &CCEdit &TF-ICON &PN-Inj. &D-Inv.\\
\hline
Clip-I Score     &88.72 &72.19 &90.19 &91.49 &\textbf{91.77} \\
DINO Score  & 0.645 & 0.534 & 0.735 & 0.743 & \textbf{0.751} \\
Clip-T Score     &26.75 &24.13 &26.67 &\textbf{26.89} &26.84 \\
Adv-Viclip Score & 0.483 & 0.322 & 0.762 & 0.747 & \textbf{0.812} \\
\end{tabular}
\caption{Quantitative Comparison in various metrics.}\label{quant}
\end{table}

\noindent \textbf{Overall Quality by Adversarial Evaluation}.
Compared to frame-wise fidelity, overall consistency is more important for a video and has a greater impact on the visual experience. Inspired by VBench~\cite{vbench}, we use Viclip~\cite{viclip} to assess the overall consistency. Due to the specific nature of the insertion task, it is inaccurate to solely calculate the distance between prompts and outputs, as the background video may already match the information in the prompt. For example, in ``Winter-Cat'', the background video inherently contains semantic information about ``\textit{cat walking}''. Even if the model fails to insert the object, the output video may still receive a high prompt-matching score.

Therefore, for each object-video pair and its optimal prompt, we designed a fake prompt, forming an adversarial library consisting of 28 prompts (e.g., \textit{``One cat walks towards the camera''} as the fake prompt and \textit{``Two cats walk towards the camera''} as the optimal one). Viclip will calculate the distance between each output and all prompts in the library. The probability value of the optimal prompt is used as the final result, namely \textit{Adversarial Viclip (Adv-Viclip)}  score.
Results are given in Tab.~\ref{quant} where D-Inv demonstrates significant advantages (0.483 $\rightarrow$ 0.812, 68\% improvement against AnyV2V). For all the prompts, discussions and more details, please refer to Appendix.

\subsection{Qualitative Evaluation}\label{sec_qual_eval}
Shown in Fig.~\ref{cat}, we provide visual examples to compare DreamInsert with baseline methods. The results of CCEdit are given in the Appendix, as it failed to insert object into the video. On the left of Fig.~\ref{cat} is the ``Cabin-Dog'', which contains a background video about a cabin with wooden floors and a burning fireplace. The object is a German Shepherd dog and the trajectory is in linear movement. All methods uses ``\textit{A german shepherd walking on the floor}'' as the text prompt. 
Although the dog in the output of AnyV2V shows motions that align with the action of ``\textit{walk}'' in the prompt, it violates the real-world logic where the dog walks into the burning fireplace. The results generated by DreamInsert-D-Inv achieve superior consistency in both spatial appearance and temporal movement, leading to reasonable environmental interaction.

On the right, ``Winter-Cat'' represents a more challenging case where the inserted cat occupies the main body of the frames and the position of the b-box remains almost unchanged, containing few movement information.
DreamInsert achieves superior visual results due to its good text-matching ability.
However,  AnyV2V and CCEdit failed to insert the object correctly, and TF-ICON was unable to maintain spatial or temporal consistency.

\subsection{User Study}\label{us}
Moreover, we provide a detailed user study, where 20 users rated 5 aspects (\textbf{Fidelity} of the inserted object, \textbf{Smoothness} of the generated object motion, \textbf{Interaction} with the background, the \textbf{Text-Matching} quality, and the overall \textbf{Quality}) on 7 videos and give a rating from 1 to 5 (ranging from failed, poor, fair, good, to excellent). For each aspect, we propose a question, and users respond to it by ratings. For example, for Fidelity, the question is: ``\textit{Is the object in the synthesized video consistent with the object in the reference image?}''. Please refer to Appendix for more details. 

Tab.~\ref{us1} shows the average score of different methods in those aspects, where DreamInsert achieves significant improvements. For overall quality, PN-Inj and D-Inv achieve 65.1\% and 86.3\% improvements compared to AnyV2V, which has the best performance among the baselines. D-Inv acquires a 16.8\% gain (3.33 $\rightarrow$ 3.89) towards PN-Inj and 78.4\% gain toward AnyV2V (2.18 $\rightarrow$ 3.89) on the reasonable \textit{interaction}. For the average rating, D-Inv achieves a 77.0\% gain (2.26 $\rightarrow$ 4.00) towards AnyV2V and a 11.4\%  gain (3.59 $\rightarrow$ 4.00) towards PN-Inj. \vspace{-0.05in}

\begin{table}[h]\footnotesize
\centering
\setlength{\abovecaptionskip}{3pt}
\setlength{\belowcaptionskip}{-13pt} 
\tabcolsep=0.13cm
\begin{tabular}{l||ccc|cc}
 & AnyV2V & CCEdit  & TF-ICON & PN-Inj. & D-Inv. \\
\hline
Fidelity      & 2.73 & 1.30 & 2.22 & 3.84 & \textbf{4.06} \\
Smoothness    & 2.11 & 1.24 & 2.18 & 3.59 & \textbf{4.07} \\
Interaction   & 2.18 & 1.21 & 1.98 & 3.33 & \textbf{3.89} \\
Text-matching & 2.14 & 1.23 & 2.35 & 3.69 & \textbf{4.04} \\
Quality       & 2.12 & 1.24 & 2.09 & 3.50 & \textbf{3.95} \\
\hline
Average       & 2.26 & 1.24 & 2.16 & 3.59 & \textbf{4.00} \\
\end{tabular}
\caption{User study comparison with the maximum score being 5.}
\label{us1}
\end{table}

\subsection{Ablation Study}\label{as}
We perform ablation studies towards the noise scale added in PN-Inj and the feature injection steps in the $2^{nd}$ stage. Results are shown in Fig.~\ref{cat_abla} and Fig.~\ref{batman}, and more studies are given in the Appendix.

\noindent \textbf{Noise scale in PN-Inj}. We discussed how the noise scale affects the performance of PN-Inj (Eqn.~\ref{pninj}). 
As shown in Fig.~\ref{cat_abla}, we set $\sigma_1$=0.4 and $\sigma_2$=0.1 as the default setting which in bold font. 
If $\sigma_1$ is set too high, $\sigma_1$=1.0 in the right-bottom, it prevents the region from generating reasonable interactions, while a too small $\sigma_1$=0.2 (right-top), will result in the diversity vanishing and the region not being able to generate motions. Setting the noise scale in the object area too high ($\sigma_2$=0.5) will cause distortion. 

\begin{figure}[!htp]
  \centering
  \setlength{\belowcaptionskip}{-8pt}
\includegraphics[width=1\linewidth]{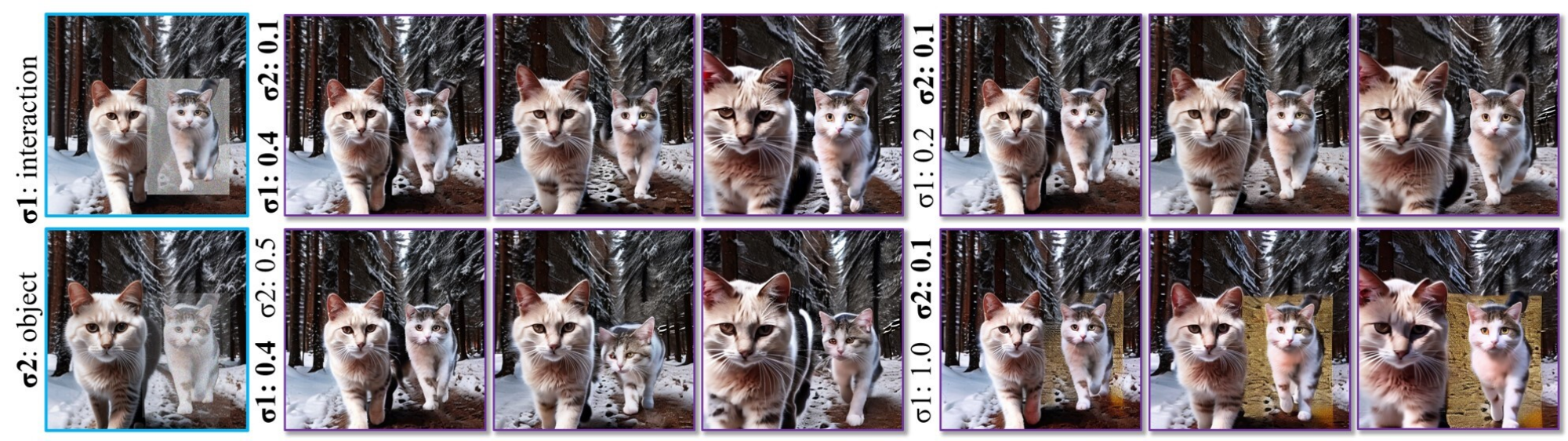}\\
  \caption{Ablation study on scale of noise added in PN-Inj, where $\sigma_1$ for the interaction area and $\sigma_2$ for the object area.  
  }
  \label{cat_abla}
\end{figure}

\noindent \textbf{Feature injection}. We discussed the impact of the injection steps in D-Inv. In Fig.~\ref{batman} of ``Library-Batman'', the numbers from left to right represent the first few steps of injection: spatial feature, spatial attention score, and temporal attention score, of the total of 50 denoising steps. Fig.~\ref{batman} shows that an early stop of feature and score injection will harm the fidelity, while more steps may result in the object vanishing or blurring. 
\vspace{-0.02in}

\begin{figure}[!htp]
  \centering
  \setlength{\belowcaptionskip}{-10pt}
 \includegraphics[width=1\linewidth]{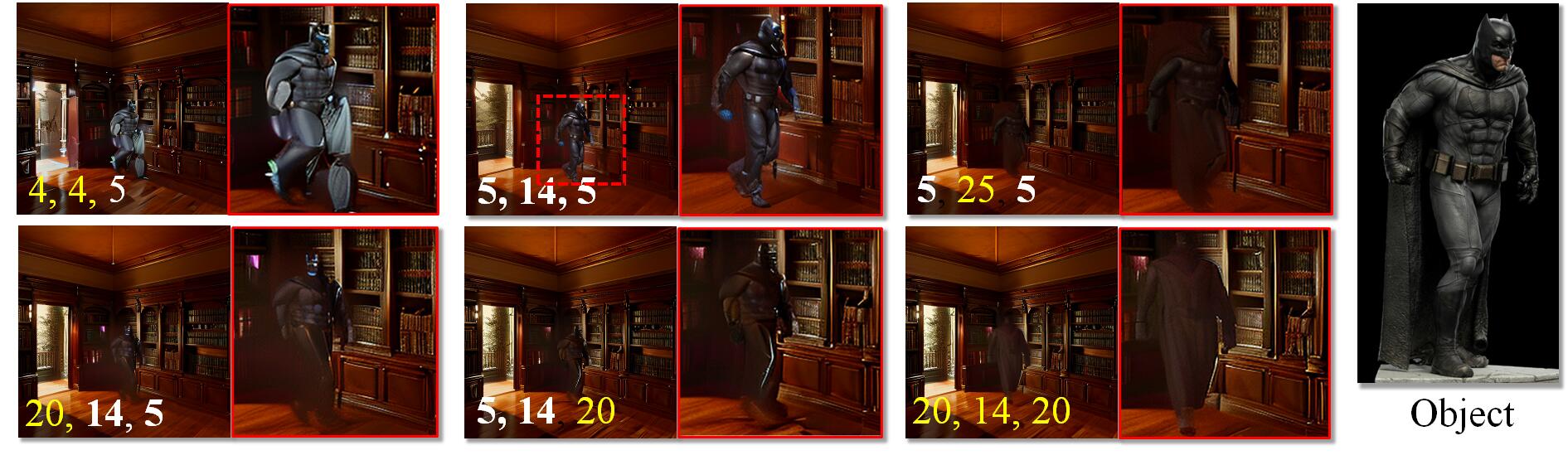}\\
  \caption{Ablation study on feature injection steps.
  }
  \label{batman}\vspace{-0.05in}
\end{figure}
\section{Conclusion}
We introduce DreamInsert, a novel framework for Image-to-Video Insertion, achieving this task in a training-free manner for the first time. Leveraging object trajectory and textual guidance, DreamInsert generates realistic motion and seamlessly integrates objects into background videos. Various experiments validate its effectiveness, marking a significant step forward in video content creation and opening new directions for future research in video synthesis.

\noindent \textbf{Limitations}.  
DreamInsert relies on I2V generative models for spatiotemporal alignment which is essential for insertion. However, current I2V models struggle with highly complex scenes like intricate limb movements, often leading to inconsistencies. As a result, DreamInsert inherits these limitations, facing challenges in handling such complex scenarios effectively.

{
\small
\bibliographystyle{unsrt}
\bibliography{main}
}
\clearpage
\clearpage
\setcounter{page}{1}
\setcounter{section}{0}
\maketitlesupplementary

\section{Experiments Detail}

\subsection{I2VIns Dataset}
To compare the performance of different methods and demonstrate the effectiveness of our approach on Image-to-Video (I2V) object insertion, we propose the \textbf{I2V} \textbf{Ins}ertion (\textbf{I2VIns}) Dataset, which containing 14 video-object pairs and each case includes a background video, an object image, and a pre-prepared trajectory.

\noindent \textbf{Background Video}.
The background videos in the I2VIns dataset include both synthesized and real videos for a better evaluation of the insertion performance. 
Each background video contains 16 frames in the shape of $512 \times 512$. 
5 of the videos are from DAVIS16~\cite{davis16}, which are ``Carround'', ``Lucia'', ``Crossing'', ``Gold-fish'', ``Dog-goose''.
The rest of the videos are synthesized videos which are generated by AnimateDiff~\cite{animatediff}, and some of them using MotionLora to control the camera motion, such as ``Hill-Horse'' and ``Winter-Cat''. 
The corresponding prompts for each background video are as follows:
\begin{enumerate}

 \item [-] Cabin-Dog: ``\textit{A rustic cabin living room with a stone fireplace and animal skins.}''

 \item [-] Coffee-Bird: ``\textit{A steaming cup of coffee on a wooden table, with the aroma and steam creating a warm, inviting atmosphere.}''

 \item [-] Forest-Horse: ``\textit{The scene transitions to a lush, green forest, with birds chirping and leaves rustling in the breeze, raw photo, beautiful shadow, hyperrealism, ultra high res, 4K.}''

 \item [-] Hill-Horse: ``\textit{Yellow wildflowers gently swaying in the breeze, rolling hills, clear blue sky, river, 8k uhd, soft lighting, high quality.}''
(MotionLora: ``PanLeft''.)

 \item [-] House-Dog: ``\textit{The scene shifts to a cozy, inviting living room, with a crackling fireplace and soft, comfortable seating. 
}''

 \item [-] Lib-Batman: ``\textit{A gothic library with dark wood shelves and candlelight.}''

 \item [-] Winter-Cat: ``\textit{A cat walks towards the camera in a snowy foresty, 8k uhd, soft lighting, high quality.}''
(MotionLora: ``Zoom-In''.)
\end{enumerate}

\begin{figure}[!h]
  \centering
\includegraphics[width=1\linewidth]{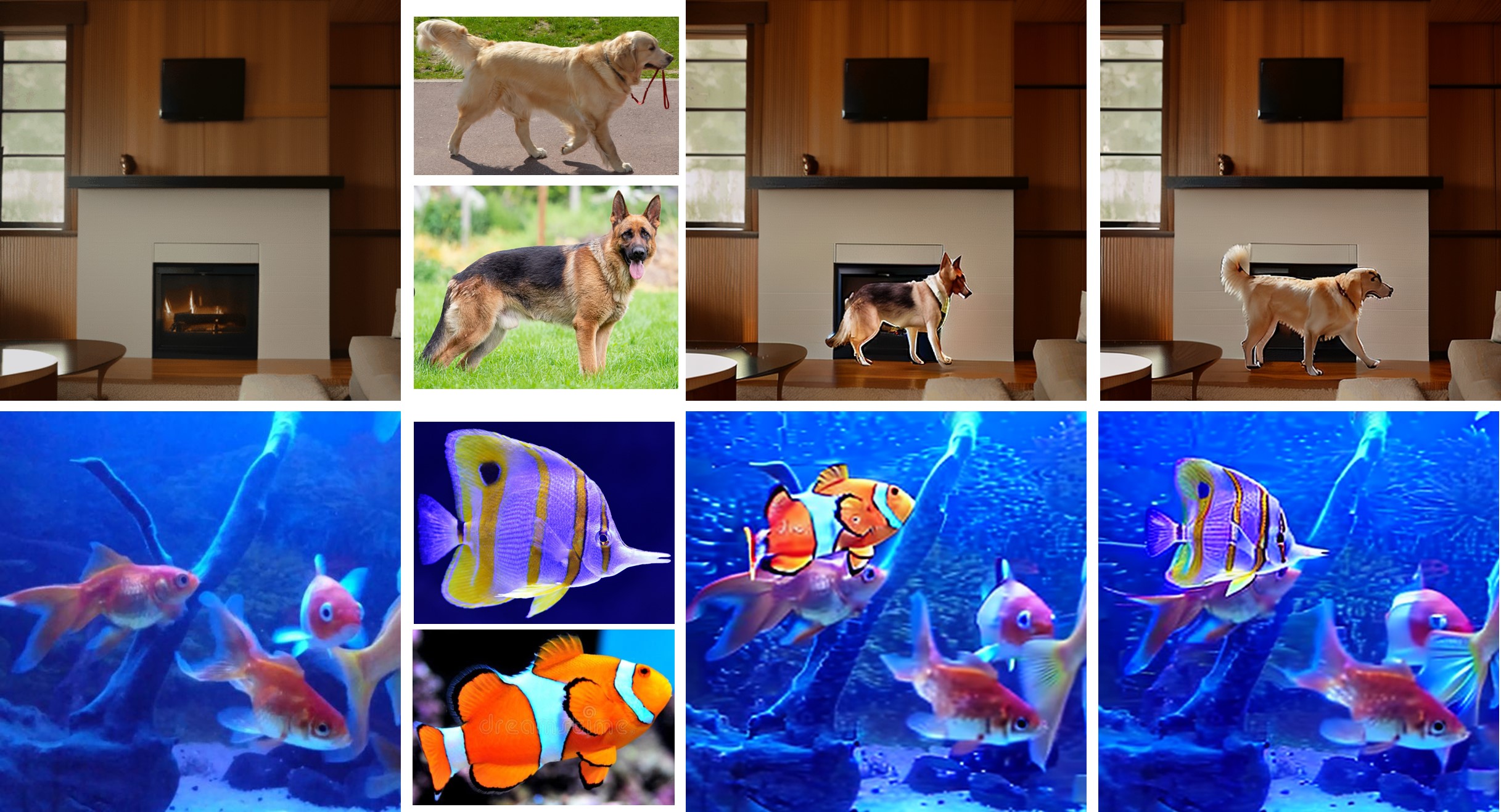}\\
  \caption{Visual examples of inserting different objects into the same scene, showing the robustness of DreamInsert.}
  \label{dog_fish}%\vspace{-0.2in}
\end{figure}

\begin{figure*}[!t]
  \centering
 \includegraphics[width=1\linewidth]{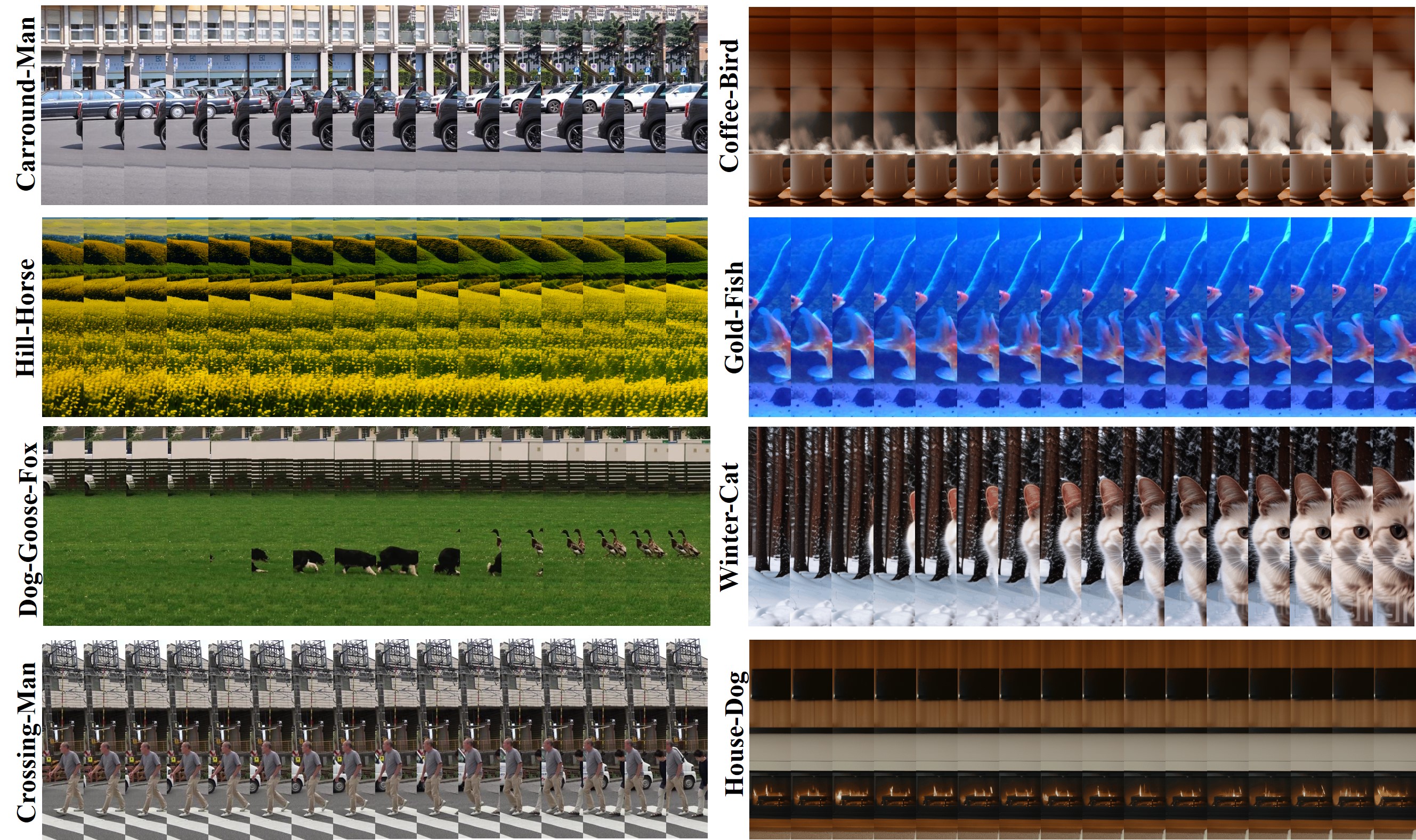}\\
  \caption{Visual examples of the background video used in I2VIns. Each clip is cut from the corresponding frame.}
  \label{bg_dynamic}%\vspace{-0.2in}
\end{figure*}

The videos cover a variety of scenes, from indoor to outdoor, showcasing the diversity of the selected cases. We demonstrate the dynamics of the background by cropping clips at the same position from eacg original frame, as shown in Fig.~\ref{bg_dynamic}. ``Cabin-Dog'' and ``Lib-Batman'' are two videos with relatively small background movement, where Cabin-Dog's firelace has a larger dynamic, creating a contrast with the surrounding static environment, while Lib-Batman's scene has a high contrast of brightness and darkness. ``Coffee-Bird'' has complex smoke variations, while Crossing Man includes the movement of people and vehicles. ``Hill-Horse'', ``Forest-Horse'' and ``Winter-Cat'' contain significant movements in the background, and accurately modeling these movements is difficult for video synthesis models.

\noindent \textbf{Object Image}. The object images in the cases cover animals and people including side and frontal perspectives. We chose them to better showcase both the motion creation after insertion and the reconstruction of interactions with the environment.

Regarding the selection of objects, unlike in 3D scene object insertion~\cite{inserf, Instruct-NeRF2NeRF, FocalDreamer, Dipir}, we did not choose still life or rigid body objects for demonstration. The main difficulty in 3D scene object insertion lies in rendering lighting and materials that correspond to the environment. In contrast, the challenges in I2V object insertion involve predicting the motion of static objects and reconstructing their environmental interactions.
To verify the robustness of model, we prepare different objects for the same scene in ``House-Dog'' and ``Gold-Fish'', resulting in 4 different cases. Visualizations are shown in Fig.~\ref{dog_fish}.

\noindent \textbf{Trajectory}.
We control the movement of the object through the trajectory, where the trajectory is generated by first giving the initial position and bounding box size, and then generating subsequent bounding boxes frame by frame according to different speeds, directions, and b-box size changes, and setting the value inside the bounding box to 1 and the value outside to 0.

\subsection{Baseline Methods}
To compare DreamInsert with existing object insertion methods, we choose three state-of-the-art content editing methods as baselines, containing training-based and training-free methods:

\noindent \textit{Training-Free Methods}: 
\begin{enumerate}
\item [-] \textbf{TF-ICON}~\cite{tficon} is an Image-to-Image (I2I) object insertion method that inserts an object from a given image into a target image using latent fusion. However, it does not perform well on realistic or natural images. We take the object image, trajectory frame, and background frame in the same timestamp as inputs and apply TF-ICON separately on each frame to obtain the synthesized result.

\item [-] \textbf{AnyV2V}~\cite{anyv2v} is an Image-to-Video (I2V) video editing method, achieves content editing using the first frame as guidance. However, AnyV2V only performs well in tasks such as style transfer or object replacement, and it is difficult for AnyV2V to predict the motion of unseen new objects. We copy the object to the target position following the trajectory to obtain the first frame, and the model uses prompt and the first frame as conditions to predict subsequent frames.
\end{enumerate}

\noindent \textit{Training-based Methods}:
\begin{enumerate}
\item [-] \textbf{CCEdit}~\cite{ccedit} (Text-to-Video, T2V). We have adopted the official default model for CCEdit to control the insertion of objects through text. However, as CCEdit was not trained on the dataset for I2V target insertion, it is completely unable to perform such task. In practice, we take background video as input and attempt to add objects and their motion through text prompts.
\end{enumerate}

\noindent We choose the above method to demonstrate the performance of different existing pipelines on this problem. However, in fact, there is currently no method to achieve the I2V object insertion task because of the challenges mentioned in the main text, namely the lack of reference motion for static objects and the absence of image-video data pairs as training data.

\begin{figure*}[!t]
  \centering
\includegraphics[width=1\linewidth]{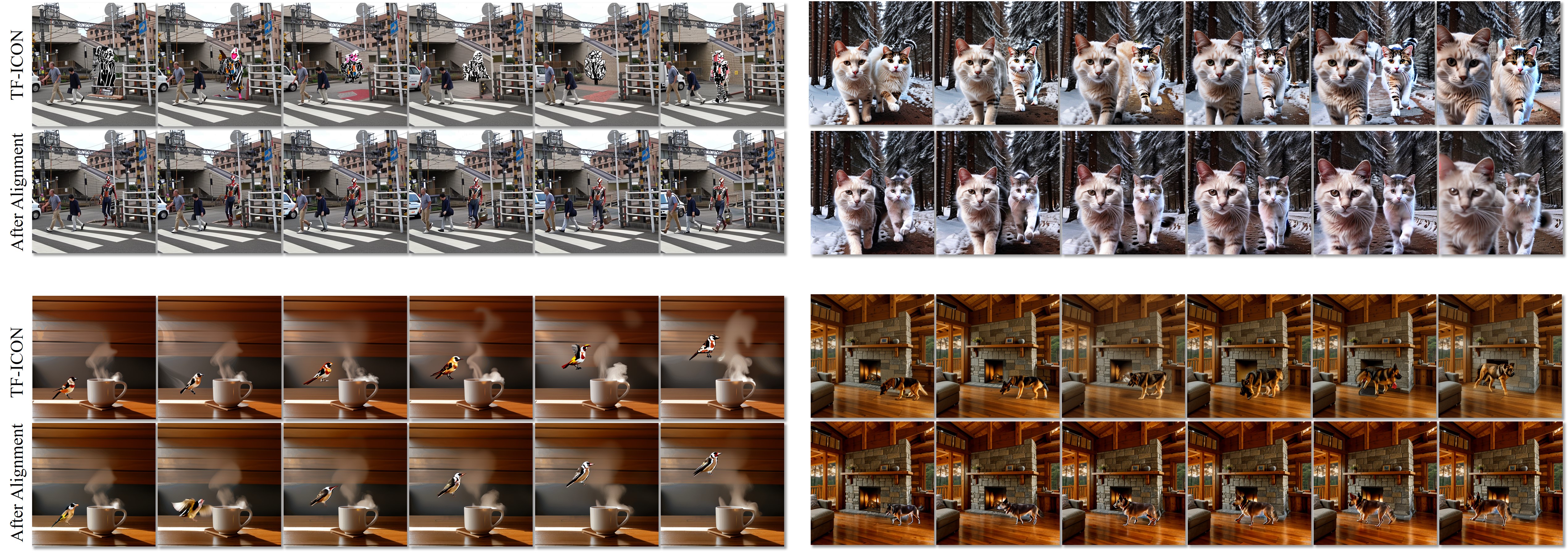}\\
 \caption{Ablation studies of motion creation in the 1st stage and spatiotemporal alignment in the 2nd stage. The top row in the left column indicates the effectiveness of motion creation in DreamInsert. Compared with the proposed motion creation, existing methods cannot generate motions for those difficult cases (Coffee-Bird and Crossing-Man), resulting in inconsistent object motions even after the alignment. 
The right-hand column reflects the effectiveness of spatial-temporal alignment in DreamInsert. For the Winter-Cat and Cabin-Dog, although there is still a certain gap between them and DreamInsert, the results of TF-ICON~\cite{tficon} are refined after the proposed alignment, indicating the effectiveness of proposed pipeline. 
 }
  \label{abla_tf}%\vspace{-0.2in}
\end{figure*}

\section{Ablation Study}
\subsection{Motion Creation}
We provide additional ablation study results to validate the effectiveness of the two stages processing in DreamInsert. We first validated the motion creation in the 1st stage. We compared the results obtained by synthesizing the latent domains (TF-ICON~\cite{tficon}), with motion created by DreamInsert. We input the former results into the model for subsequent alignment, and the results are shown in Fig.~\ref{abla_tf}. 

Comparing the results in the left column of Fig.~\ref{abla_tf} with those in Fig.~\ref{supp_bird} and Fig.~\ref{supp_crossing}, it can be seen that our proposed pixel-domain fusion can generate more diverse motions while preserving semantic consistency, resulting in better performance, while the output of TF-ICON, despite alignment, still cannot obtain motions consistent with the object.

Comparing the results in the right column of Fig.~\ref{abla_tf} with those in Fig.~\ref{supp_cat} and Fig.~\ref{supp_dog}, it demonstrates the effectiveness of our proposed spatiotemporal alignment in the 2nd stage. For motion generated by latent-domain fusion that lacks consistency, alignment via the knowledge from I2V generation model can improve its consistency with objects. 

\subsection{Object Robustness}
We design following experiment to verify the robustness of the model on different objects, as shown in Fig.~\ref{supp_fish1} and Fig.~\ref{supp_fish2}. Among them, we kept the prompt, trajectory, and background video unchanged, and two completely different objects were inserted. Among them, Gold-Fish1 uses a clownfish with a significantly different appearance from the background, while Gold-Fish2 uses a blue fish that is very similar to the background. From the visualizations, it can be seen that due to the presence of fish of the same type as the new object in the background, it is difficult for the baseline method to correctly distinguish them, resulting in incorrect insertion results. Regardless of domain differences, DreamInsert achieved reasonable insertion results. D-Inv shows better robustness as it correctly models the spatial relationship between objects and background while keeping the fidelity.

\section{Quantitative Comparison}
We offer various quantitative comparisons among three aspects of \textbf{Fidelity}, \textbf{Text-Matching} and \textbf{Overall Quailty}, to compare the performance between different methods on the I2VIns Dataset. We use Clip-I, Clip-T score~\cite{clipscore}, DINO Score~\cite{dino} and Adv-Viclip score~\cite{viclip} as metrics. DreamInsert in both the PN-Inj and D-Inv settings achieves comparable results, showing its effectiveness.

\subsection{Fidelity Comparison using Clip-I and DINO Score}
\noindent \textbf{Clip-I Score}.
We calculated the average Clip Image (Clip-I) score by comparing generated frames to the copied sequence to evaluate the faithfulness between the output and the original background and object. Maintaining the fidelity of the background is also important in I2V insertion. The results are presented in Tab.~\ref{clipi}. 

\begin{table}[!h]\footnotesize
\centering
\setlength{\abovecaptionskip}{3pt}
\setlength{\belowcaptionskip}{-10pt} 
\tabcolsep=0.15cm
%\resizebox{!}{1.3cm}{\centering
\begin{tabular}{l||ccc|cc}
Clip-I Score  & AnyV2V &CCEdit &TF-ICON &PN-Inj. &D-Inv.\\
\midrule[1.2pt]
Cabin-Dog    &96.29 &89.81 &95.25 &95.45 &\textbf{96.43}  \\
Carround-Man	&86.77	&70.74	&\textbf{87.64	}&86.27	&85.67 \\
Coffee-Bird  &91.23 &77.81 &92.77 &89.31 &\textbf{95.11}  \\
Crossing-Man &77.71 &59.62 &76.26 &\textbf{91.97} &91.93  \\
Dog-Goose-Fox	&81.85	&67.20	&\textbf{89.05	}&79.94	&85.92 \\
Forest-Horse &93.80 &73.76 &92.80 &\textbf{93.99} &93.77 \\
Gold-Fish1	&91.80	&80.64	&89.04	&\textbf{94.52	}&91.17 \\
Gold-Fish2	&92.32	&81.64	&87.49	&\textbf{92.90}	&91.58 \\
Hill-Horse   &84.75 &67.66 &93.96 &93.35 &\textbf{96.09} \\
House-Dog1	&89.95	&64.15	&90.25	&\textbf{92.48}	&92.27 \\
House-Dog2	&95.64	&70.29	&94.45	&\textbf{94.98}	&93.55 \\
Lib-Batman   &84.00 &68.24 &90.86 &\textbf{94.85} &92.80 \\ 
Lucia-Dog	&82.37	&63.44	&\textbf{86.92	}&83.71	&82.86 \\
Winter-Cat   &93.62 &75.73 &95.99 &\textbf{97.15} &95.69 \\   
\hline
Average		&88.72	&72.19	&90.19	&91.49	&\textbf{91.77}
\end{tabular}
\caption{Comparison of frame fidelity using Clip-Image Score.}\label{clipi}
\end{table}

\noindent \textbf{DINO Score}.
In order to more accurately evaluate the fidelity of the inserted object, we follow the VBench~\cite{vbench} setting, using DINO score~\cite{dino} as metric.
We calculate the DINO score between the reference image and the bounding-box area of each synthesized frame. Differ from Clip-I score which measure the distance between two frames, DINO score calculates the object area between the object image and predicted frame, can more accurately evaluate the fidelity of the object itself.

\begin{table}[h]\footnotesize
\centering
\setlength{\abovecaptionskip}{3pt}
\setlength{\belowcaptionskip}{-10pt} 
\tabcolsep=0.15cm
\begin{tabular}{l||ccc|cc}
DINO Score & AnyV2V & CCEdit & TF-ICON & PN-Inj. & D-Inv. \\
    \midrule[1.2pt]
Cabin-Dog    & 0.591 & 0.486 & \textbf{0.737} & 0.692 & 0.700 \\
Carround-Man &0.682	&0.601	&0.690	&0.657	&\textbf{0.693} \\
Coffee-Bird  & 0.567 & 0.561 & \textbf{0.680} & 0.616 & 0.662 \\
Crossing-Man & 0.542 & 0.513 & 0.601 & \textbf{0.700} & 0.692 \\
Dog-Goose-Fox	&0.708	&0.508	&0.742	&0.753	&\textbf{0.764} \\
Forest-Horse & 0.667 & 0.534 & 0.839 & \textbf{0.846} & 0.813 \\
Gold-Fish1	&0.843	&0.715	&0.646	&\textbf{0.852}	&0.813 \\
Gold-Fish2	&0.807	&0.720	&0.899	&0.906	&\textbf{0.913} \\
House-Dog1	&0.575	&0.470	&\textbf{0.695}	&0.605	&0.645 \\
House-Dog2	&0.682	&0.491	&0.747	&\textbf{0.750}	&0.742 \\
Hill-Horse   & 0.618 & 0.526 & 0.811 & 0.765 & \textbf{0.815} \\
Lib-Batman   & 0.543 & 0.529 & 0.645 & \textbf{0.694} & 0.655 \\
Lucia-Dog	&0.639	&0.514	&0.732	&0.725	&\textbf{0.755} \\
Winter-Cat   & 0.578 & 0.587 & 0.828 & 0.840 & \textbf{0.860} \\
\hline
Average & 0.645 & 0.534 & 0.735 & 0.743 & \textbf{0.751} \\
\end{tabular}
\caption{Comparison of object fidelity using DINO score.}
\label{dino_score}
\end{table}

\noindent \textbf{Results}.
As shown in Tab.~\ref{clipi} and Tab.~\ref{dino_score}, D-Inv achieves the best results (91.77 of Clip-I and 0.751 of Dino). DreamInsert achieved outstanding performance in maintaining the fidelity of both the background and object, with both settings outperforming baseline methods.

\subsection{Text-Matching Comparison using Clip-T Score}
We calculated the average Clip Text (Clip-T) score between each generated frame of each model's output and the input text prompt, thereby gauging the accuracy of the created motions instructed by the text. The results are given in Table~\ref{clipt}. PN-Inj achieves the best results, with D-Inv coming in second. The score of all methods on the Clip-I score is not high, because the prompt used in I2V insertion is intended to control the motion of the object and therefore does not include background description. The Clip-T score calculates the distance between each frame and the prompt separately, which introduces a certain degree of error. Nevertheless, under the same prompt, DreamInsert still achieves the best results, demonstrating that the generated motion is more in line with the prompt's instructions.

\begin{table}[!h]\footnotesize
\centering
\setlength{\abovecaptionskip}{3pt}
\setlength{\belowcaptionskip}{-12pt} 
       \tabcolsep=0.15cm
 \begin{tabular}{l||ccc|cc}
Clip-T Score   & AnyV2V &TF-ICON &CCEdit &PN-Inj. &D-Inv. \\
    \midrule[1.2pt]

Cabin-Dog    &17.64 &\textbf{23.83} &18.25 &21.04 &21.14 \\
Carround-Man	&24.91	&18.97	&\textbf{25.82}	&25.42	&25.01   \\
Coffee-Bird  &20.93 &25.70 &\textbf{28.39} &21.79 &25.10 \\
Crossing-Man &24.13 &24.82 &25.08 &\textbf{24.86} &24.44 \\
Dog-Goose-Fox	&33.20	&\textbf{33.75}	&32.82	&33.32	&33.51   \\
Forest-Horse &29.05 &\textbf{31.18} &27.63 &30.38 &30.37 \\
Gold-Fish1	&28.22	&\textbf{29.20}	&27.86	&28.26	&28.17   \\
Gold-Fish2	&28.26	&27.16	&\textbf{28.83}	&28.35	&28.11   \\  
Hill-Horse   &22.81 &\textbf{30.04} &22.28 &29.85 &29.56 \\
House-Dog1	&21.32	&17.58	&18.96	&\textbf{21.66}	&21.62   \\
House-Dog2	&24.67	&18.42	&24.28	&24.44	&\textbf{25.16}   \\
Lib-Batman   &20.06 &25.22 &20.11 &\textbf{25.28} &25.26 \\ 
Lucia-Dog	&26.72	&23.88	&\textbf{28.14}	&26.79	&26.34   \\
Winter-Cat   &28.95 &\textbf{30.12} &29.42 &29.95 &29.71 \\     
 \hline
Average		&26.75	&24.13	&26.67	&\textbf{26.89}	&26.84   \\
  \end{tabular}
   \caption{Comparison of text-matching performance using Clip-Text Score.}\label{clipt}
\end{table}

\begin{table}[!t]\footnotesize
\centering
\setlength{\abovecaptionskip}{3pt}
\setlength{\belowcaptionskip}{-10pt} 
\tabcolsep=0.15cm
\begin{tabular}{l||l}
Case & Prompt \quad (\textit{Top}: Optimal, \textit{Bottom}: Fake)\\
    \midrule[1.2pt]

Coffee-Bird &A bird flies slowly, head upward. \\
  &A bird walks slowly to the top, head upward. \\ \hline
Cabin-Dog & A german shepherd walks to the left of the cabin.\\
  &A german shepherd stands on the floor. \\ \hline
Crossing-Man &Three man walk through\\ 
    &Two man walk through.\\ \hline
Forest-Horse &A horse walking slowly to the left. \\
    &A horse walking around \\ \hline
Hill-Horse &Horse running\\ 
  &Horse eating grass.\\ \hline
Lib-Batman &Batman walks into the room. \\
   &Batman flies into the room. \\ \hline
Winter-Cat &Two cats walk towards the camera. \\
    &One cat walks towards the camera. \\ \hline
House-Dog1 	&A german shepherd walks to the right of room. \\
	&A german shepherd walks in the forest. \\ \hline
House-Dog2 &A golden retriever walks to the right of the room. \\
&A golden retriever walks to the left of the room. \\ \hline
Dog-Goose-Fox &Fox chasing the goose. \\
            &Fox chasing the goose and dog. \\ \hline
Carround-Man &A man walks next to a black car. \\
       &A man driving the black car \\ \hline
Lucia-Dog  &A woman walking the dog. \\
      &A woman walking and dog stop. \\ \hline
Gold-Fish1 &A clownfish swimming with other three fishes. \\
        &Three clownfish are swimming. \\ \hline
Gold-Fish2 &A blue fish swimming with other three fishes. \\
        &A blue fish swimming with other two fishes. \\ \hline
\end{tabular}
\caption{Adversarial Prompt list with optimal prompt on the top and fake one on the bottom.}\label{viclip_prompt}
\end{table}

\subsection{Overall Quality Comparison using Adversarial Viclip Score}

We use a pretrained large-scale video-text model, Viclip~\cite{viclip}, to evaluate the overall consistency. Viclip is built upon Clip and consists of a video encoder (ViT) and a text encoder. Both modules are initialized from the corresponding Clip components. It incorporates both contrastive learning and masked modeling techniques, allowing for efficient learning of transferable video-language representations. We use Viclip to evaluate the insertion quality and propose the \textit{Adversarial Viclip (Adv-Viclip) Score} .

\begin{table}[!t]\footnotesize
\centering
\setlength{\abovecaptionskip}{3pt}
\setlength{\belowcaptionskip}{-10pt} 
\tabcolsep=0.15cm
\begin{tabular}{l||ccc|cc}
Adv-Viclip & AnyV2V & CCEdit & TF-ICON & PN-Inj. & D-Inv. \\
    \midrule[1.2pt]
Cabin-Dog & 0.960 & 0.801 & 0.990 & 0.991 & \textbf{0.992} \\
Carround-Man & 0.981 & 0.803 & 0.994 & 0.994 & \textbf{0.995} \\
Coffee-Bird & 0.525 & 0.740 & \textbf{0.762} & 0.745 & 0.743 \\
Crossing-Man & 0.150 & 0.418 & 0.379 & 0.274 & \textbf{0.443} \\
Dog-Goose-Fox & 0.771 & 0.838 & \textbf{0.930} & 0.793 & 0.860 \\
Forest-Horse & 0.827 & 0.001 & 0.841 & 0.804 & \textbf{0.900} \\
Gold-Fish1 & 0.573 & 0.211 & 0.003 & \textbf{0.683} & 0.582 \\
Gold-Fish2 & 0.530 & 0.017 & \textbf{0.685} & 0.393 & 0.590 \\
Hill-Horse & 0.686 & 0.001 & 0.959 & \textbf{0.976} & 0.953 \\
House-Dog1 & 0.576 & 0.001 & 0.478 & 0.675 & \textbf{0.697} \\
House-Dog2 & 0.524 & 0.043 & \textbf{0.643} & 0.490 & 0.426 \\
Lib-Batman & 0.602 & 0.335 & 0.664 & 0.727 & \textbf{0.835} \\
Lucia-Dog & 0.584 & 0.552 & 0.601 & 0.633 & \textbf{0.670} \\
Winter-Cat & 0.303 & 0.682 & 0.980 & \textbf{0.993} & 0.987 \\
\hline
Average & 0.613 & 0.388 & 0.707 & 0.726 & \textbf{0.763} \\
\end{tabular}
\caption{Comparison of overall quality using Adv-Viclip score.}
\label{viclip}
\end{table}

In detail, we provide a list of optimal prompts, $\mathbb{L}^{\text{opt}}$, each corresponding to a case in the I2VIns dataset, which describes the ideal synthesis effect for that case. For the synthesis results obtained by different models, Viclip calculates the distance between the synthesis results and all the prompts in $\mathbb{L}^{\text{opt}}$ and outputs the probability corresponding to each prompt. We take the probability value of the ground truth prompt as the result to evaluate the performance.

Due to the unique nature of the insertion task, calculating only the probability between the output video and the optimal prompt is insufficient to evaluate the performance. This is because the differences between each case can be quite significant, meaning that even if the insertion quality is not good enough, Viclip can still find the optimal prompt as the closest match with a high probability value. Therefore, we design a set of fake prompts, $\mathbb{L}^{\text{fak}}$. Compared to the optimal prompt, each fake prompt in $\mathbb{L}^{\text{fak}}$ corresponds to one in $\mathbb{L}^{\text{opt}}$, describing the same case in I2VIns. However, the prompts in $\mathbb{L}^{\text{fak}}$ describe failure situations of the case, emphasizing the difference between the good and fake ones. Viclip will search for the prompt that best matches the generated results from the combined list $\mathbb{L}^{\text{adv}} = \mathbb{L}^{\text{opt}} \cup \mathbb{L}^{\text{fak}}$, thereby providing a more accurate evaluation. 
For the final probability value which measured on $\mathbb{L}^{\text{adv}}$, we refer to it as Adversarial Viclip (Adv-Viclip) Score.
All the results of different models using Adv-Viclip score are shown in Table~\ref{viclip}, and the prompt list $\mathbb{L}^{\text{adv}}$ is presented in Table~\ref{viclip_prompt}.

\section{User Study}
We randomly selected 7 videos for the user study. 20 users evaluated the insertion performance of these 7 cases, each containing insertion results from 5 different methods. The evaluation indicators are divided into five aspects, namely Fidelity, Motion Smoothness, Environment Interaction, Text-matching, and Overall Quality. For each aspect, we pose a question and users rate it on a scale of 1 to 5 (Failed, Poor, Fair, Good, and Perfect). The five questions are as follows:
\begin{enumerate}
\item [-] \textit{Fidelity}: Is the object in the synthesized video consistent with the object in the reference image?

\item [-] \textit{Motion Smoothness}: Is the object's motion in the video reasonable?

\item [-] \textit{Interaction}: Is the integration and interaction of new objects with the surrounding background harmonious in composite videos?

\item [-] \textit{Text-matching}: Does the synthesized video match the textual description provided?

\item [-] \textit{Overall Quality}: What is the overall performance of the result?
\end{enumerate}

\begin{table}[!t]\footnotesize
    \centering
\setlength{\belowcaptionskip}{-5pt} 
   \tabcolsep=0.13cm
\begin{tabular}{l||ccc|cc}
Videos & AnyV2V & CCEdit & TF-ICON & PN-Inj. & D-Inv. \\
\toprule
Cabin-Dog & 2.99 & 1.35  & 1.96 & 3.40 &\textbf{ 3.83} \\
Coffee-bird & 2.24 & 1.31 & 2.89 & 2.69 & \textbf{3.91} \\
Crossing-Man & 1.71 & 1.41 & 1.45 & \textbf{3.96} & 3.79 \\
Forest-Horse & 2.36 & 1.13  & 2.49 & 3.64 & \textbf{4.08} \\
Hill-Horse & 2.25 & 1.08 & 2.09 & 3.20 & \textbf{3.93} \\
Lib-Batman & 2.01 & 1.12  & 1.99 & 4.01 & \textbf{4.31} \\
Winter-Cat & 2.24 & 1.32  & 2.28 & \textbf{4.28} & 4.17 \\
\hline 
Average    & 2.26 & 1.24  & 2.16 & 3.59 & \textbf{4.00} \\
    \end{tabular}
    \caption{User study comparison on different videos.}
    \label{us2}
\end{table}

\noindent \textbf{Results}. The evaluations of each video are given in Tab.~\ref{us2}.
The average score of DreamInsert showed significant competitiveness, with PN-Inj receiving a score higher than 3 (``Fair'') compared to existing methods with a score close to 2 (i.e. evaluation of ``Poor''), while D-Inv received an evaluation score 4, close to ``Good''. Especially for the case of real videos like ``Crossing-Man'', the scores for baseline methods did not reach ``Poor'', while DreamInsert in two settings received ratings close to ``Good''.

\section{Case Study}
\begin{figure*}[!t]
  \centering
\includegraphics[width=1\linewidth]{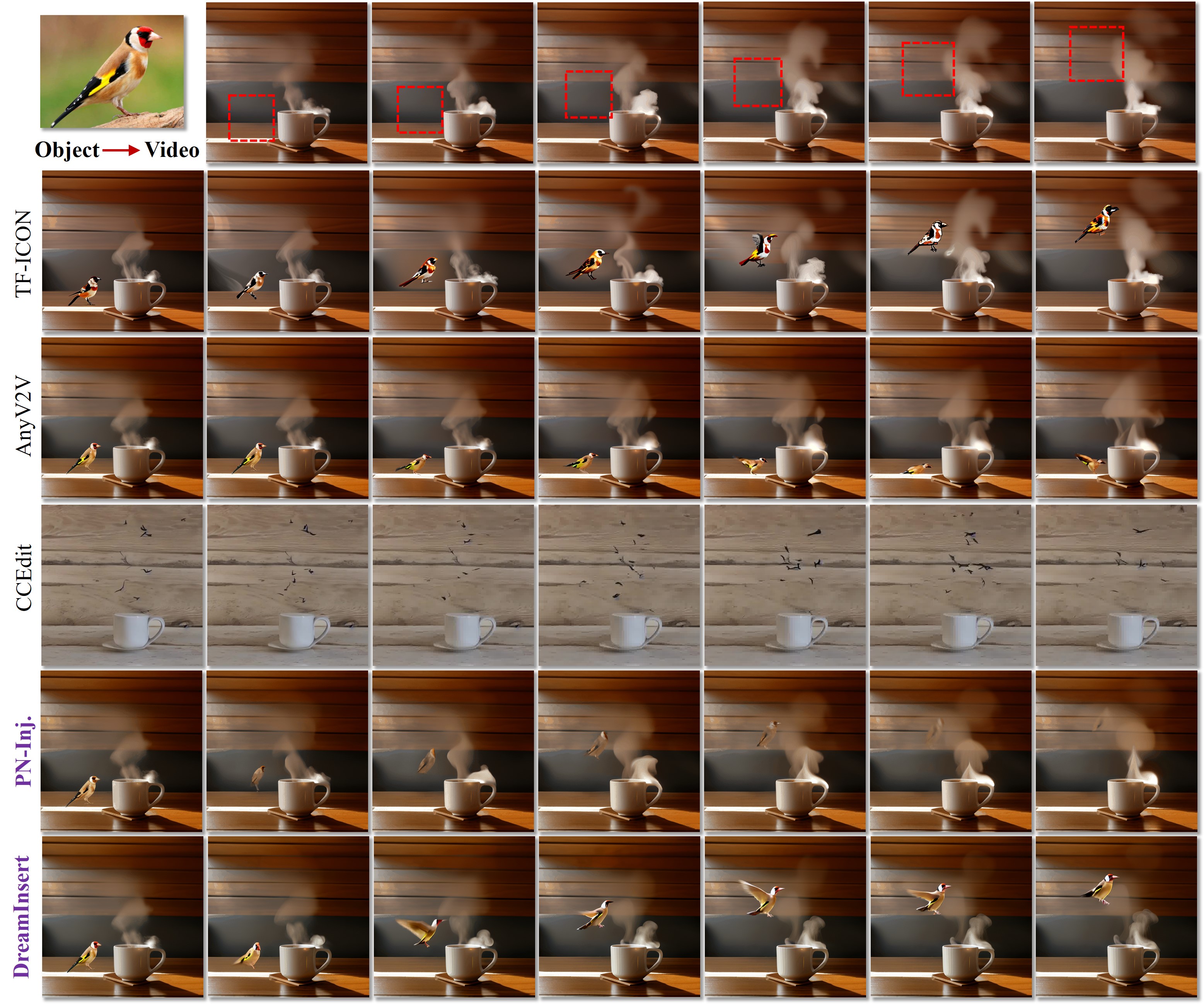}\\
 \caption{Visualized comparison on the Coffee-Bird. 
 The Coffee-Bird is the most difficult scene because the bird in the object image lacks the conditions to fly at all, and even its wings are barely recognizable. Under the PN-Inj setting, it is also challenging to reconstruct non-existent flying actions using alignment in the second stage. Only by first performing motion creation and then aligning can a coherent and consistent bird flight video be generated, enabling the insertion of objects into a new scene. It is worth mentioning that, in the setting of a bird flying next to a coffee cup, it is nearly impossible to find relevant real videos as training samples. This indicates the novelty of DreamInsert, which can achieve zero-shot object insertion. (Text prompt: \textit{A bird flies slowly, head upward}.)}
  \label{supp_bird}
\end{figure*}

\begin{figure*}[!t]
  \centering
\includegraphics[width=1\linewidth]{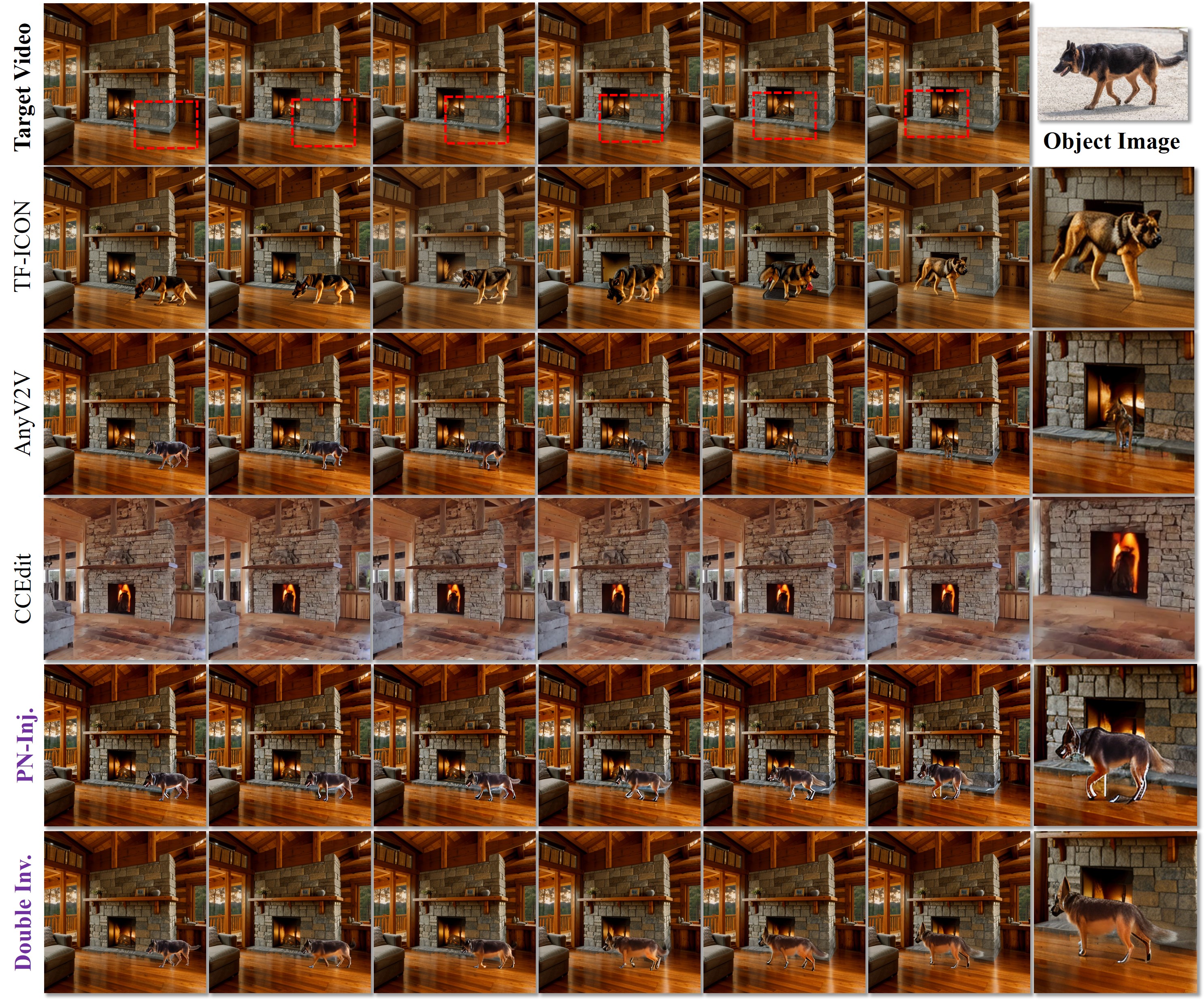}\\ \caption{Visualized comparison on Cabin-Dog.
Cabin-Dog is a relatively easy case, where both PN-Inj and D-Inv successfully generated coherent actions. However,D-Inv maintained better spatiotemporal coherence. TF-ICON~\cite{tficon} cannot understand the relationship between actions, scenes, and trajectories. Although AnyV2V~\cite{anyv2v} generates coherent actions, these actions do not conform to objective laws and cannot be precisely controlled. (Text prompt: \textit{A german shepherd walks on the floor.}.)
}
  \label{supp_dog}
\end{figure*}

\begin{figure*}[!t]
  \centering
\includegraphics[width=0.95\linewidth]{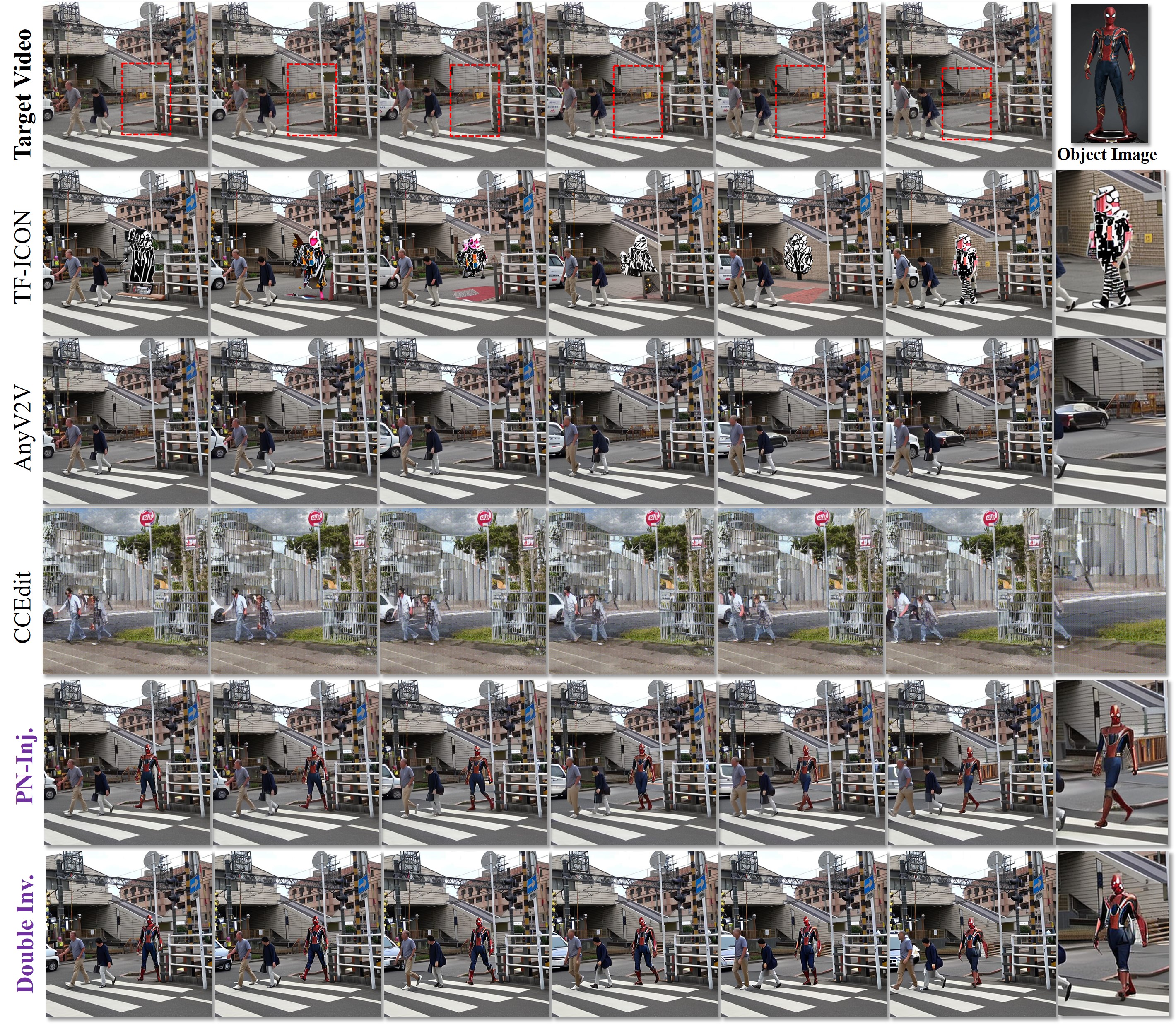}\\
 \caption{Visualized comparison on Crossing-Man.
 The main challenge of Crossing-Man lies not only in the fact that its background (bg) video is sourced from real videos (``Crossing'') in DAVIS16~\cite{davis16}, which are rich in details, but also in the fact that, due to the presence of other objects of the same category in the bg video, accurately controlling different objects within the same category poses a challenge for all models. DreamInsert achieved good results in both settings. The object vanishes completely in the second frame of AnyV2V~\cite{anyv2v}'s generation. TF-ICON~\cite{tficon} fails to even synthesize the object with the bg video, which underscores the effectiveness of our proposed pixel-level noise perturbation. In such scenarios, achieving precise synthesis and control of the object and bg frame in the latent domain, while maintaining spatiotemporal consistency across different frames, is almost impossible. (Text prompt: \textit{A man walks through.}.)
 }
\label{supp_crossing}
\end{figure*}

\begin{figure*}[!t]
  \centering
\includegraphics[width=1\linewidth]{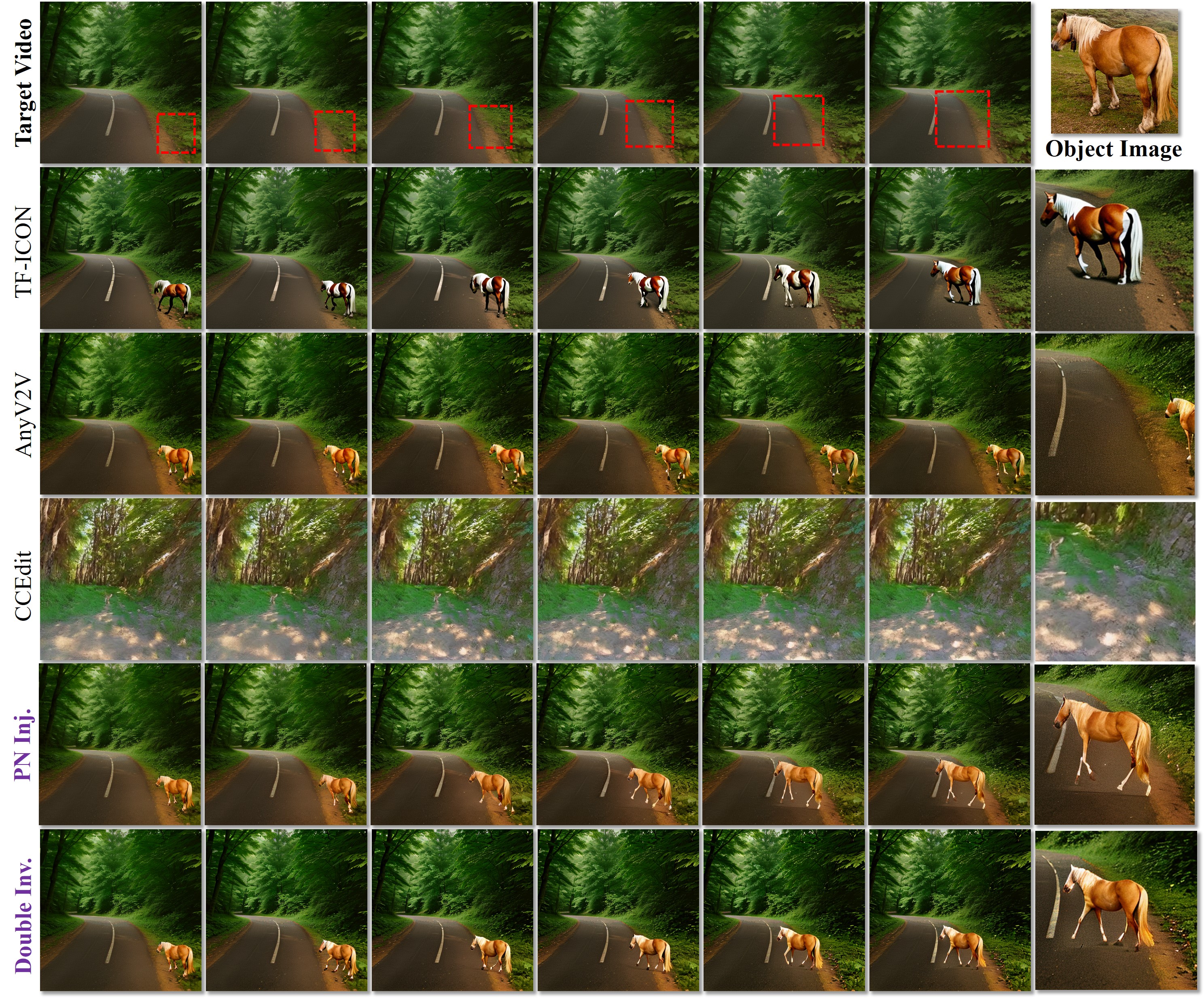}\\
 \caption{Visualized comparison on Forest-Horse.
 As mentioned in the main text, the major difficulty of Forest-Horse lies in the fact that the position and size of the bounding boxes in the trajectory change over time. Therefore, it is quite difficult to maintain the spatial consistency of objects, as the number and position of pixels they occupy are constantly changing. TF-ICON~\cite{tficon} fails to control the spatial coherence of objects in the latent domain, while AnyV2V~\cite{anyv2v} cannot control the movement of horses effectively, failing to generate coherent motion. When the bounding box size changes, PN-Inj struggles to balance the object and background in the interaction area. D-Inv, however, generates a coherent and consistent object motion. (Text prompt: \textit{A horse walking slowly to the left}.)
}
 \label{supp_forest}
\end{figure*}

\begin{figure*}[!t]
  \centering
\includegraphics[width=1\linewidth]{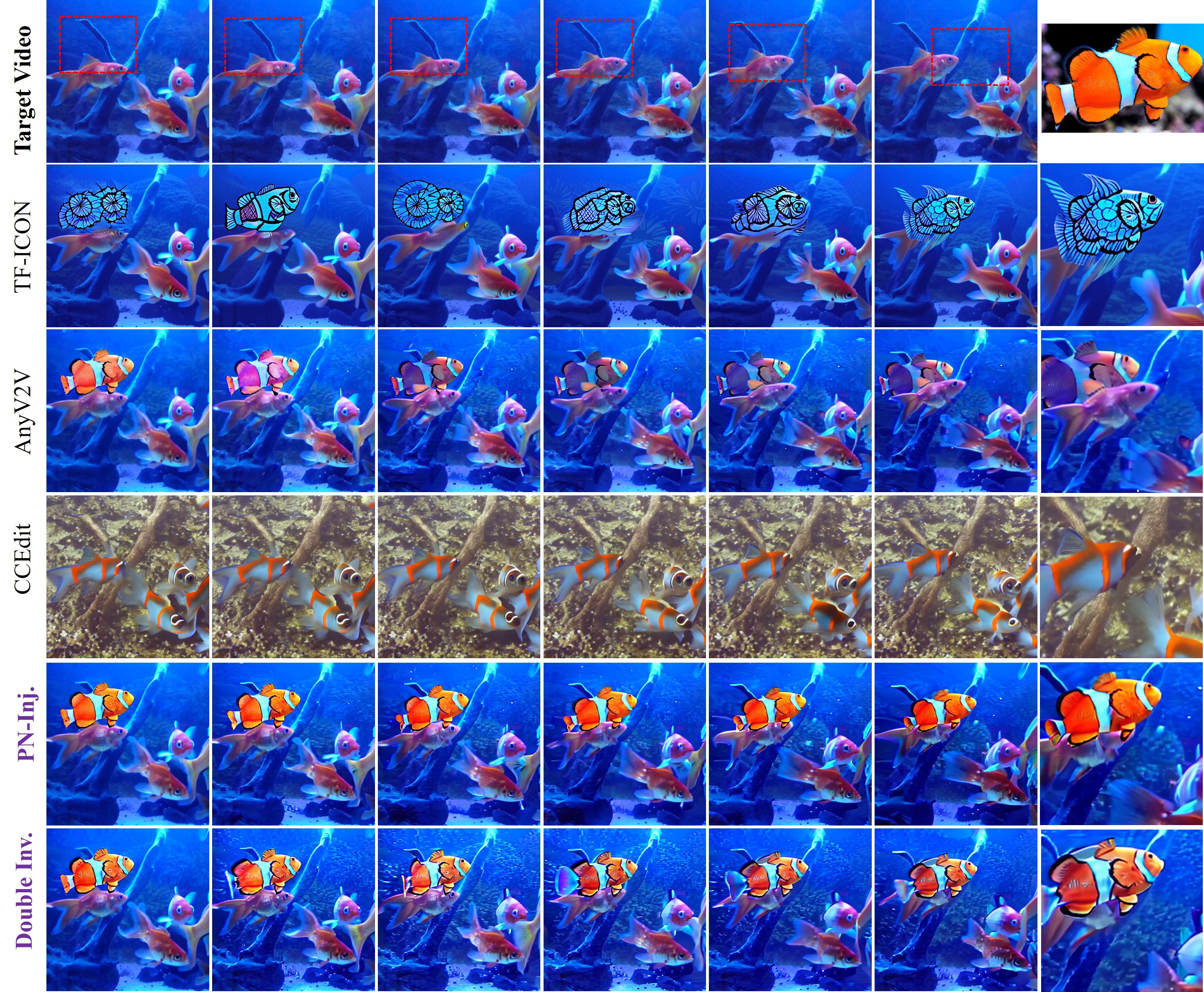}\\
 \caption{Visualized comparison on Gold-Fish1.
The background video of Gold-Fish is from Davis16~\cite{davis16}, which depicts the scene of fish swimming underwater. In Gold-Fish1, we inserted a clownfish. The difficulty of this case lies in whether the model can reasonably maintain the appearance of the clownfish. AnyV2V~\cite{anyv2v} cannot reasonably maintain the appearance of the object, and the color of the clownfish in the first few frames has noticeably faded. When two fish were stacked, TFICON~\cite{tficon} can not understand the new object of  clownfish and instead fused its appearance features with the background fish, generating an incorrect image. CCEdit~\cite{ccedit} is unable to insert the object. Compared to PN-Inj, D-Inv not only maintains the fidelity of the object, but also better simulates environmental lighting and color, achieving the best results
 (Text prompt: \textit{Fish Swimming}.)
}
 \label{supp_fish1}
\end{figure*}

\begin{figure*}[!t]
  \centering
\includegraphics[width=1\linewidth]{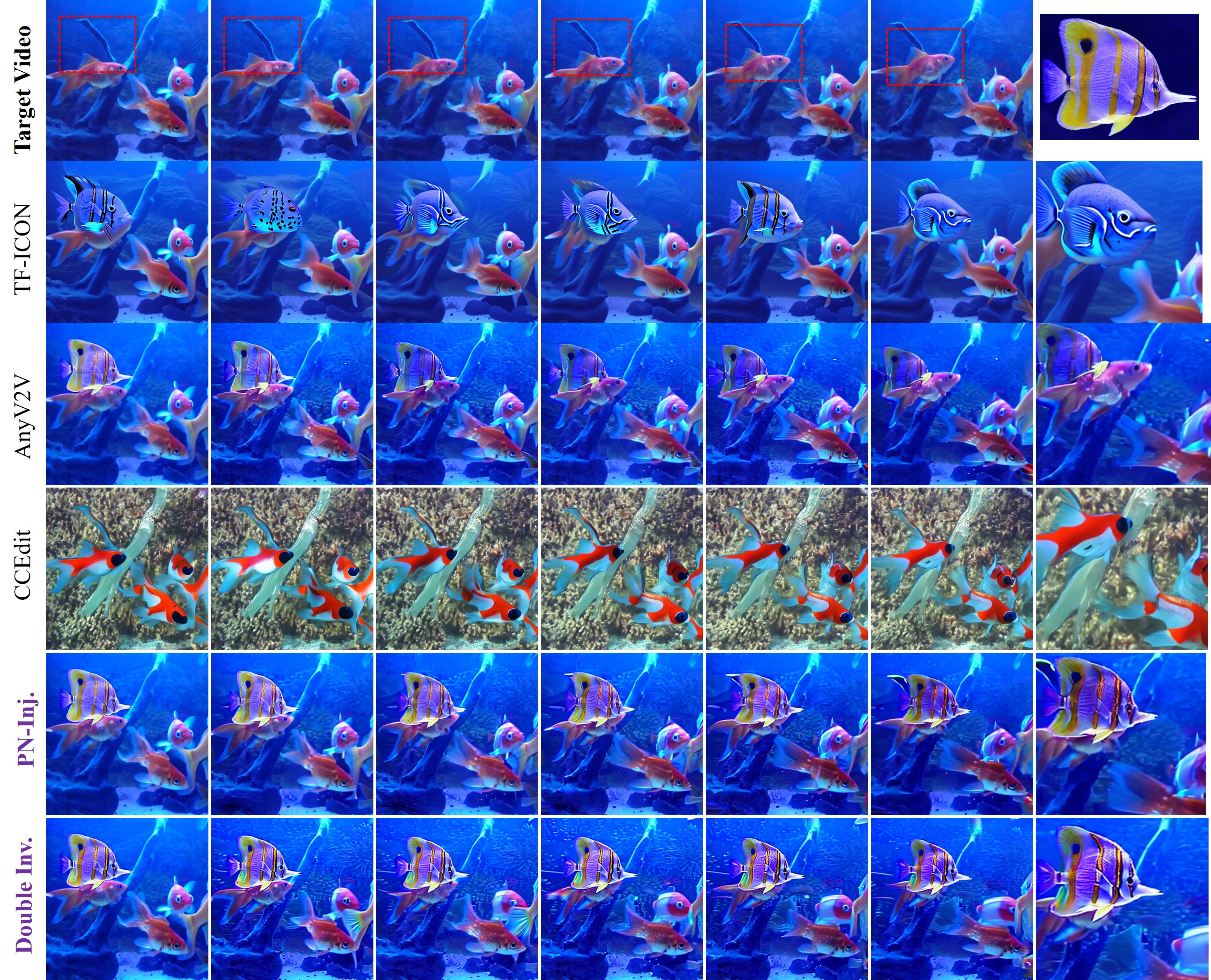}\\
 \caption{Visualized comparison on Gold-Fish2.
 In Gold-Fish2, we validated the ability to insert different objects in the same scene. We kept the prompt, trajectory, and background video unchanged, but changed the type of object. In this case, inserting a blue fish into the background is very difficult because the object's color is similar to the background, making it easy for the model to confuse it with the environment. AnyV2V~\cite{anyv2v} cannot correctly distinguish objects, resulting in incorrect fusion of blue fish and background fish, and changing the appearance of the background fish without generating reasonable interaction between the object and the background. TFICON~\cite{tficon} has once again fused two fish, and CCEdit~\cite{ccedit} is unable to insert objects. In the results of PN Inj, there were artifacts and distortions in the subsequent frame environment, while D-Inv achieved the best results, correctly modeling the spatial relationship between object fish and background fish while ensuring fidelity.
 (Text prompt: \textit{Fish Swimming}.)
}
 \label{supp_fish2}
\end{figure*}

\begin{figure*}[!t]
  \centering
\includegraphics[width=1\linewidth]{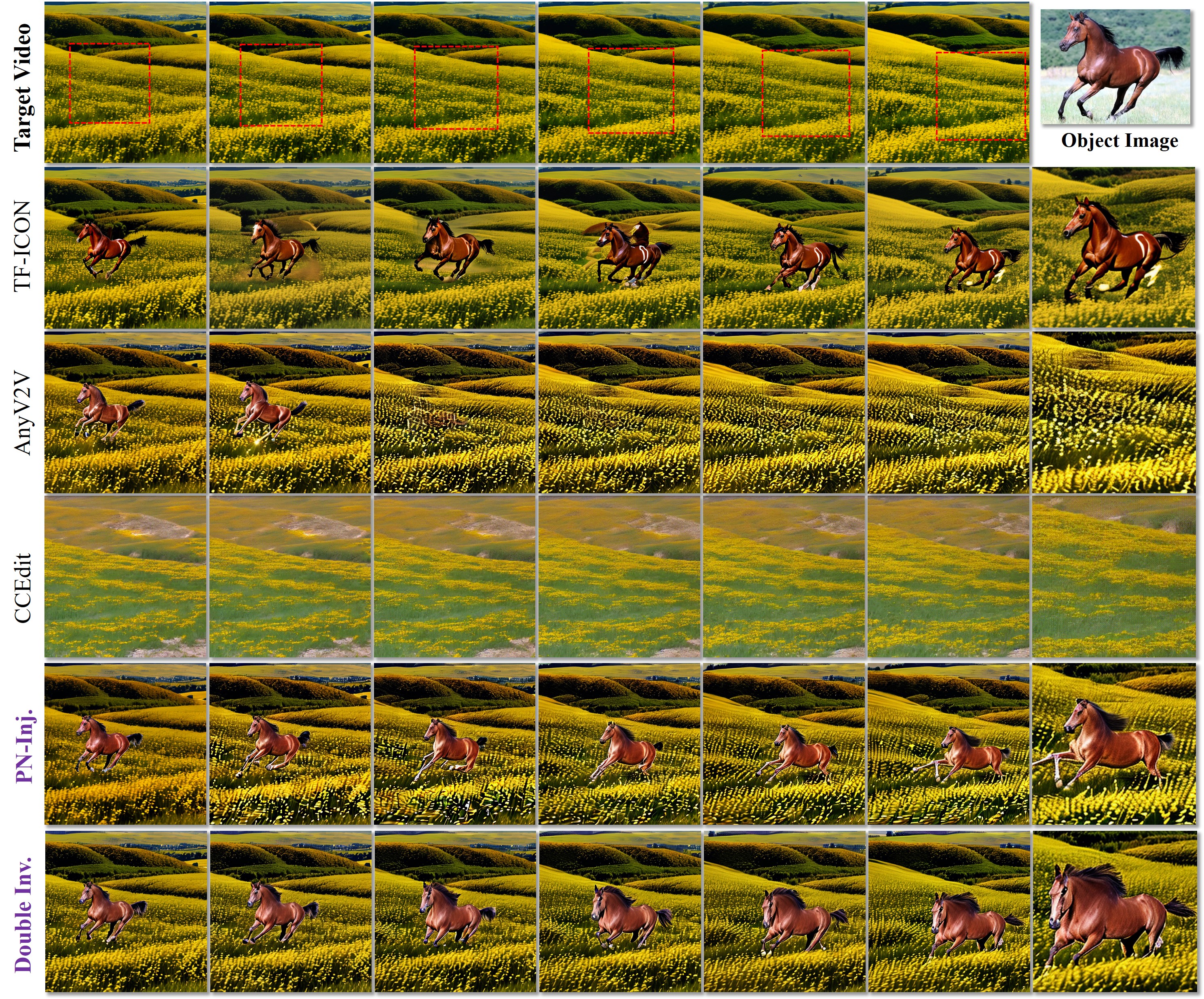}\\
 \caption{Visualized comparison on Hill-Horse.
 Also, it is very difficult to maintain consistency during the large movements of both object and background scene. The results of TF-ICON~\cite{tficon} are spatially and temporally incoherent. Due to the significant dynamics of the background, the object vanishes after the first five frames of AnyV2V~\cite{anyv2v}. PN-Inj struggles to maintain the relationship between legs and body due to the large dynamics of the scene and the intense object motion. D-Inv generates coherent and consistent object synthesis results, achieving the insertion of moving objects in large dynamic scenes.
 (Text prompt: \textit{Horse running}.)
 }
  \label{supp_hill}
\end{figure*}

\begin{figure*}[!t]
  \centering
\includegraphics[width=1\linewidth]{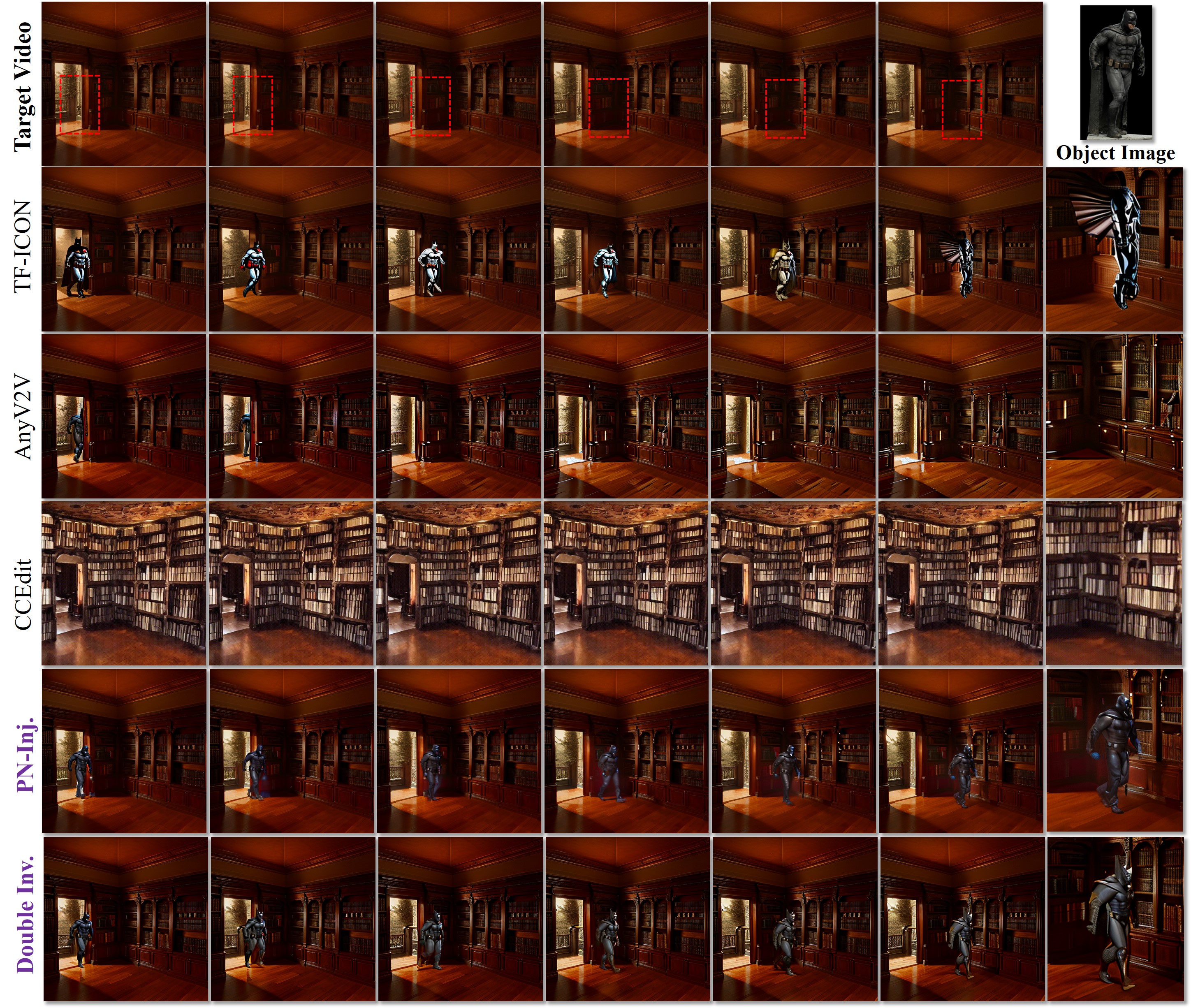}\\
 \caption{Visualized comparison on Lib-Batman.
 The difficulty of Lib-Batman lies in the fact that both objects and scenes are dim and dark, making it difficult to maintain spatial coherence of objects. TF-ICON~\cite{tficon} cannot maintain the consistency of objects, and AnyV2V~\cite{anyv2v} objects disappear in a short period of time. In this scenario, PN-Inj achieved better results because it maintained the consistency of the object subject. And based on the setting of motion creation, the object was reconstructed with bias.
 (Text prompt: \textit{Man walking inside}.)
 }
  \label{supp_lib}
\end{figure*}

\begin{figure*}[!t]
  \centering
\includegraphics[width=0.95\linewidth]{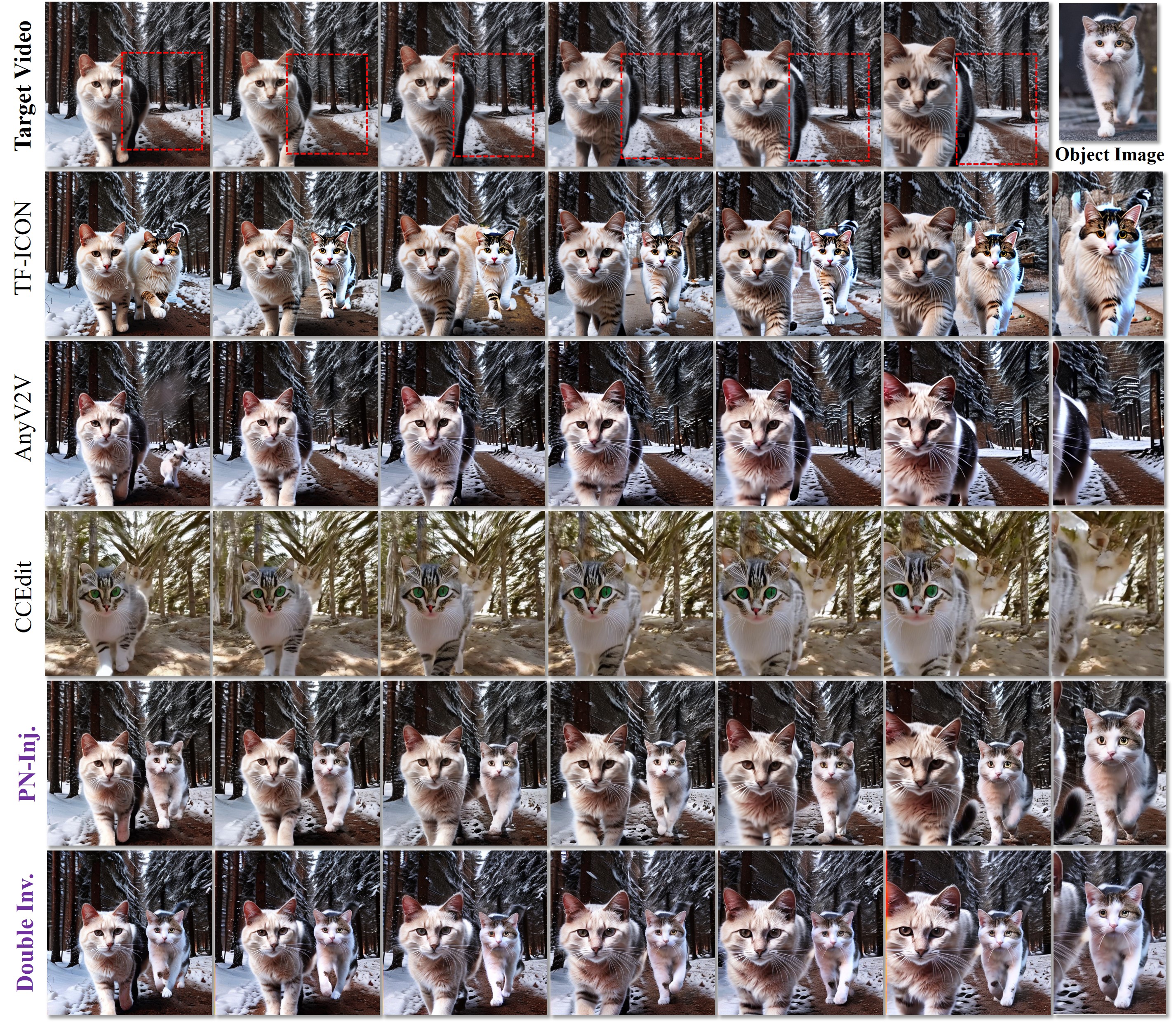}\\
 \caption{Visualized comparison on Winter-Cat.
 Winter-Cat is also a very difficult example, and its difficulty lies in: 1. The inserted object occupies a large proportion of the frame and requires consistency in its facial details; 2. The background image has significant dynamic (Zoom-In); 3. The category of the inserted object is the same as that of the original object. TF-ICON~\cite{tficon} inserted different cats frame by frame, and the cats in AnyV2V~\cite{anyv2v} disappeared within 3 frames. PN-Inj generated better insertion results, perfectly preserving the original characteristic of the cat's face and producing coherent walking movements. 
 (Text prompt: \textit{Cat walk towards the camera}.)
 }
  \label{supp_cat}
\end{figure*}

In order to better demonstrate the insertion effect of DreamInsert and compare it with baseline models, we provide several case studies. 
The comparison results are shown in~\Crefrange{supp_bird}{supp_cat} where detailed discussion of each case is presented in the caption. 
The top row is the background video, the red bounding box is the trajectory, and the top-right is the object image. The other rows are results obtained from different baselines, where ``\textcolor{violet}{PN-Inj.}'' represents the results obtained by performing pixel-level noise injection and alignment, while ``\textcolor{violet}{D-Inv.}'' is the default setting of DreamInsert, representing the results obtained by using motion creation and alignment. We presented the 2nd, 4th, 7th, 10th, 13th, and 16th frames of the output results (unless otherwise specified) of different methods for a fair comparison. In the rightmost column, we present the details of the synthesized object in the 16th frame, specifically to compare the spatial and temporal consistency among different methods.

Both the PN-Inj and D-Inv pipeline demonstrate certain effectiveness. While PN-Inj performs better for objects with little change in subject (such as the human body in Fig.~\ref{supp_crossing} and Fig.~\ref{supp_lib}, or objects in the frontal view of Fig.~\ref{supp_cat}, etc.), as pixel-level noise injection does not significantly alter the spatial features and appearance of the object. The D-Inv pipeline shows its advantages in scenarios with significant changes in object appearance (horse in Fig.~\ref{supp_hill}), where new actions that do not exist originally need to be added (e.g., bird flying in Fig.~\ref{supp_bird}), or where the bounding box in the trajectory undergoes size changes and the interaction area is complex and varied (Fig.~\ref{supp_forest}).

\begin{figure}[!h]
  \centering
\includegraphics[width=1\linewidth]{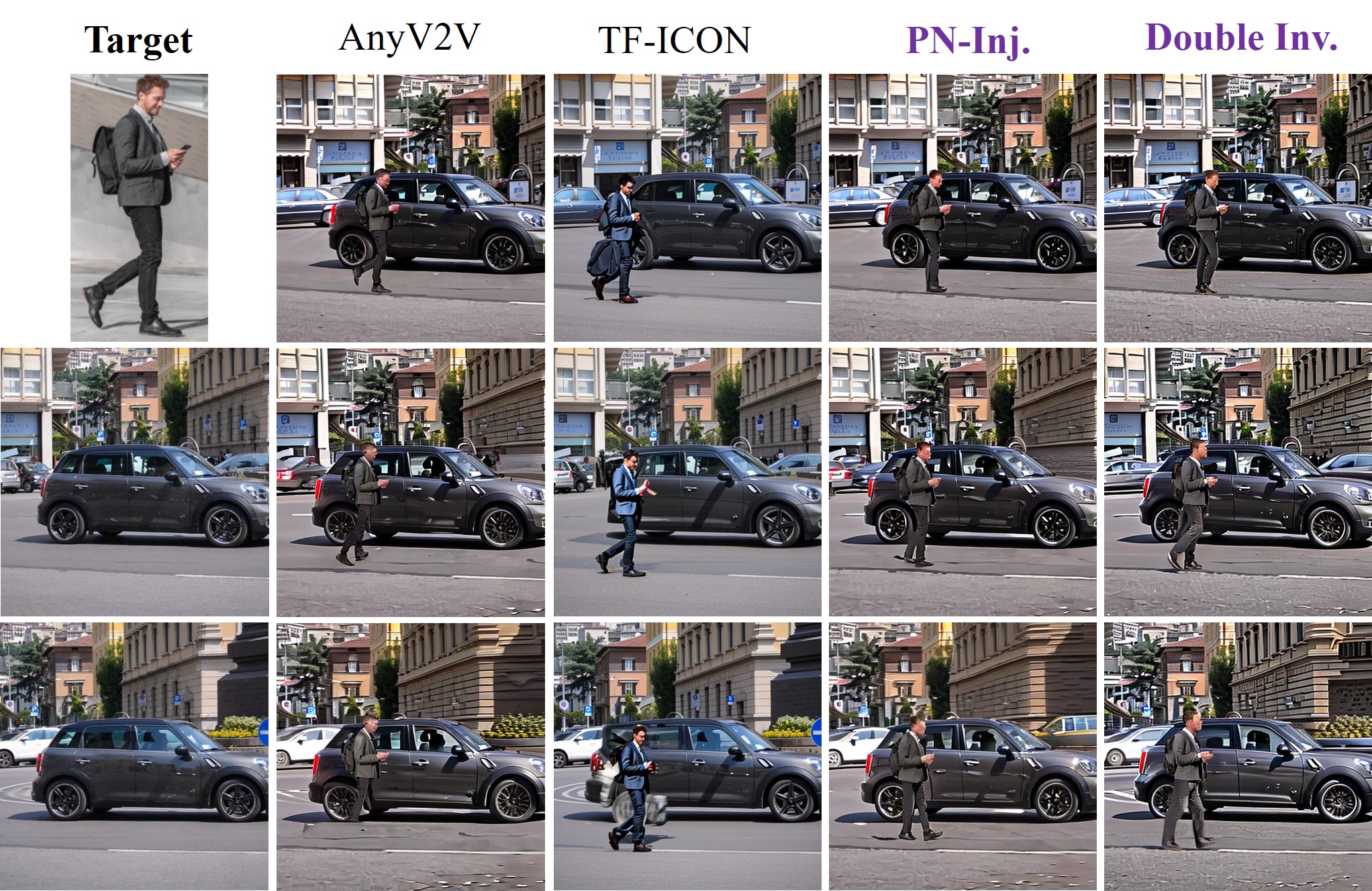}\\
 \caption{Blurring issue, using Carround-Man as a example.}
  \label{car_man}
\end{figure}

\noindent \textbf{Failure Case}
Currently, all methods have poor insertion performance when the pixels of the object are limited. 
Fig.~\ref{car_man} shows a failure case where the model attempts to insert a person into the Carround video from DAVIS16~\cite{davis16}. However, due to the small number of pixels occupied by the human face in the object image, the model finds it difficult to extract its spatial features. Especially after 8 $\times$ downsampling by the VAE in LDM~\cite{ldm}, it is difficult for the model to reconstruct the face. All of the methods have shown blurred predictions. However, due to reasonable modeling of the motion, DreamInsert still achieved comparable results.

\section{Limitations}
Currently, our method has achieved a preliminary breakthrough in I2V object insertion for specific scenarios, enabling the creation of coherent dynamic videos of certain durations. However, we face three major limitations: Firstly, the generation quality for human faces and bodies is moderate, due to humans' heightened sensitivity to the authenticity of facial and bodily features~\cite{uv}, making it challenging to maintain consistency in real videos. Secondly, synthesizing long-term video sequences is difficult because the motion states of the static object become more complex and harder to predict them over extended periods in consistency. Thirdly, accurately preserving background details during object-environment interactions remains a significant challenge for all image and video editing and synthesis algorithms. We plan to delve deeper into these issues in our future research endeavors.

\section{Social Impacts}
Our original intention is to provide possibilities for film and television production~\cite{ccdm}, or to assist in psychotherapy~\cite{seeme}, post-traumatic stress disorder (PTSD) treatment~\cite{TreatingTT}, etc., to reproduce the scenes in people's memories, or to achieve the integration of imagined scenes. Current methods for object insertion in realistic videos struggle to produce coherent synthesized videos, whereas our method is capable of generating short-term videos with consistent motion, thereby opening up new creative possibilities for such video synthesis and editing tasks and serving as a catalyst for future research. However, malicious parties may use this technology to create false news or harmful video content to mislead viewers. This is a prevalent problem that also arises in other generative model methodologies or content manipulation techniques. ongoing research in generative modeling, particularly personalized generative priors, must persist in examining and reconfirming these issues.

\end{document}